\documentclass{article}

\usepackage{microtype}
\usepackage{graphicx}
\usepackage{booktabs} 

\usepackage{hyperref}

\usepackage[accepted]{icml2024}

\usepackage{amsmath}
\usepackage{amssymb}
\usepackage{mathtools}
\usepackage{amsthm}
\usepackage{wasysym}

\usepackage[capitalize,noabbrev]{cleveref}

\theoremstyle{plain}

\theoremstyle{definition}

\theoremstyle{remark}

\usepackage[textsize=tiny]{todonotes}

\def\ie{\emph{i.e.}}

\def\vs{\emph{vs.}}

\DeclareUnicodeCharacter{2212}{-}
\newlength\savewidth\newcommand\shline{\noalign{\global\savewidth\arrayrulewidth
  \global\arrayrulewidth 1pt}\hline\noalign{\global\arrayrulewidth\savewidth}}
\usepackage{makecell}
\usepackage{multirow}
\usepackage{pifont}
\newcommand{\cmark}{\ding{51}}%
\newcommand{\xmark}{\ding{55}}%
\usepackage{float}                  
\usepackage{subfig}
\usepackage{marvosym}

\begin{document}

\twocolumn[
\icmltitle{$\gemini$GeminiFusion: Efficient Pixel-wise Multimodal Fusion for Vision Transformer}



\icmlsetsymbol{equal}{*}

\begin{icmlauthorlist}
\icmlauthor{Ding Jia}{equal,sch}
\icmlauthor{Jianyuan Guo}{equal,usyd}
\icmlauthor{Kai Han}{comp}
\icmlauthor{Han Wu}{usyd}
\icmlauthor{Chao Zhang}{sch}
\icmlauthor{Chang Xu\textsuperscript{\Letter}}{usyd}
\icmlauthor{Xinghao Chen\textsuperscript{\Letter}}{comp}
\end{icmlauthorlist}

\begin{center}
jiading@stu.pku.edu.cn; \{jianyuan.guo,han.wu\}@sydney.edu.au; kai.han@huawei.com; c.zhang@pku.edu.cn
\end{center}

\icmlaffiliation{usyd}{The University of Sydney.}
\icmlaffiliation{comp}{Huawei Noah's Ark Lab.}
\icmlaffiliation{sch}{Peking University.}

\icmlcorrespondingauthor{Chang Xu}{c.xu@sydney.edu.au}
\icmlcorrespondingauthor{Xinghao Chen}{xinghao.chen@huawei.com}

\icmlkeywords{Machine Learning, ICML}

\vskip 0.3in
]



\printAffiliationsAndNotice{\icmlEqualContribution} 

\begin{abstract}
Cross-modal transformers have demonstrated superiority in various vision tasks by effectively integrating different modalities. This paper first critiques prior token exchange methods which replace less informative tokens with inter-modal features, and demonstrate exchange based methods underperform cross-attention mechanisms, while the computational demand of the latter inevitably restricts its use with longer sequences.
To surmount the computational challenges, we propose \emph{GeminiFusion}, a pixel-wise fusion approach that capitalizes on aligned cross-modal representations. \emph{GeminiFusion} elegantly combines intra-modal and inter-modal attentions, dynamically integrating complementary information across modalities. 
We employ a layer-adaptive noise to adaptively control their interplay on a per-layer basis, thereby achieving a harmonized fusion process. 
Notably, \emph{GeminiFusion} maintains linear complexity with respect to the number of input tokens, ensuring this multimodal framework operates with efficiency comparable to unimodal networks.
Comprehensive evaluations across multimodal image-to-image translation, $3$D object detection and arbitrary-modal semantic segmentation tasks, including RGB, depth, LiDAR, event data, etc. demonstrate the superior performance of our \emph{GeminiFusion} against leading-edge techniques. The PyTorch code is available \href{https://github.com/JiaDingCN/GeminiFusion}{here}.

\end{abstract}

\section{Introduction}

\begin{figure*}[t]
\centering
\subfloat{\includegraphics[height=34mm]{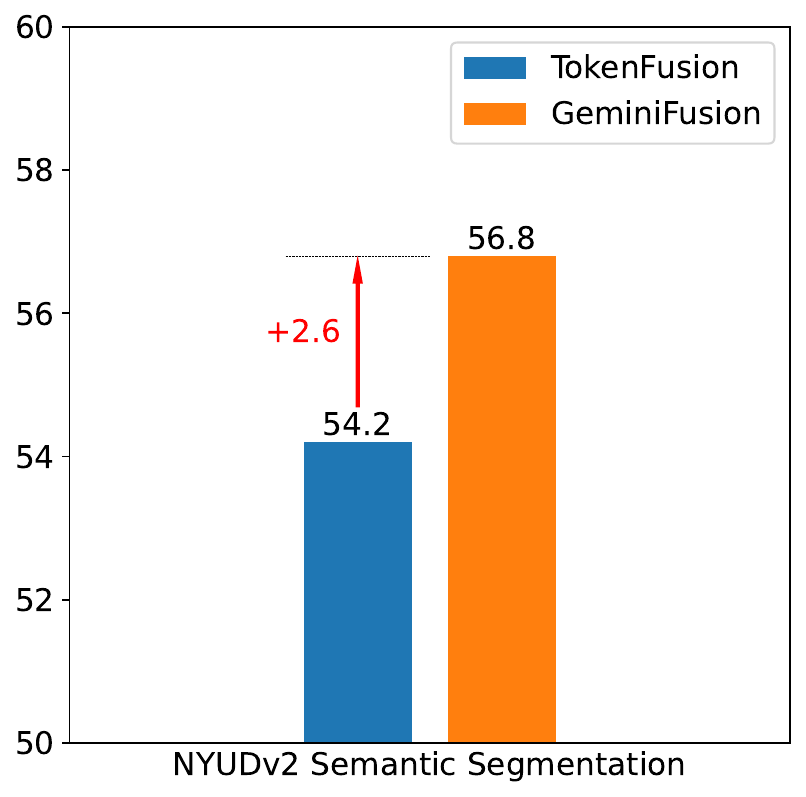}}
\subfloat{\includegraphics[height=34mm]{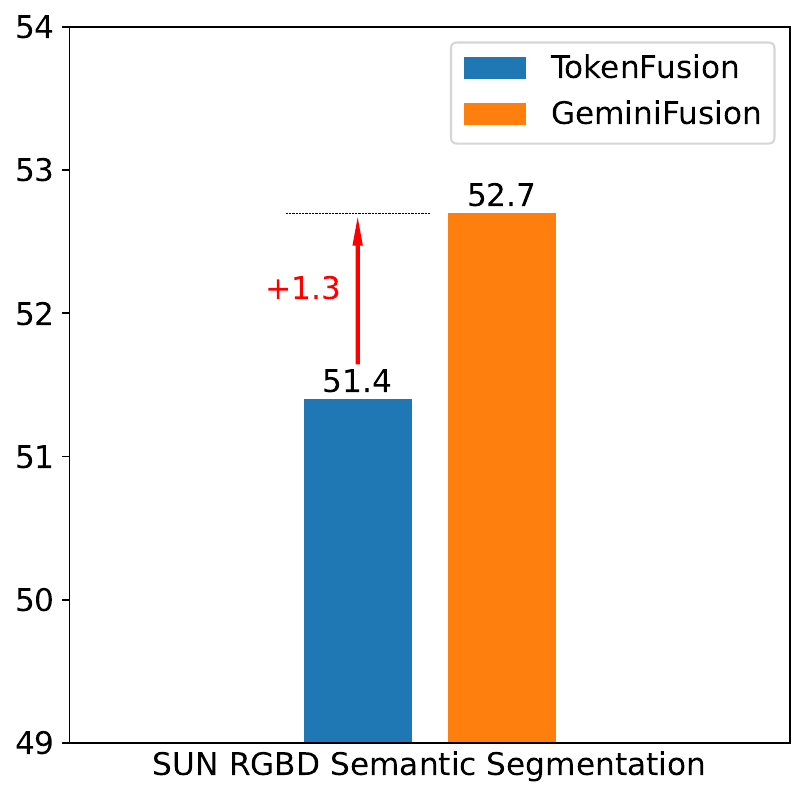}} 
\subfloat{\includegraphics[height=34mm]{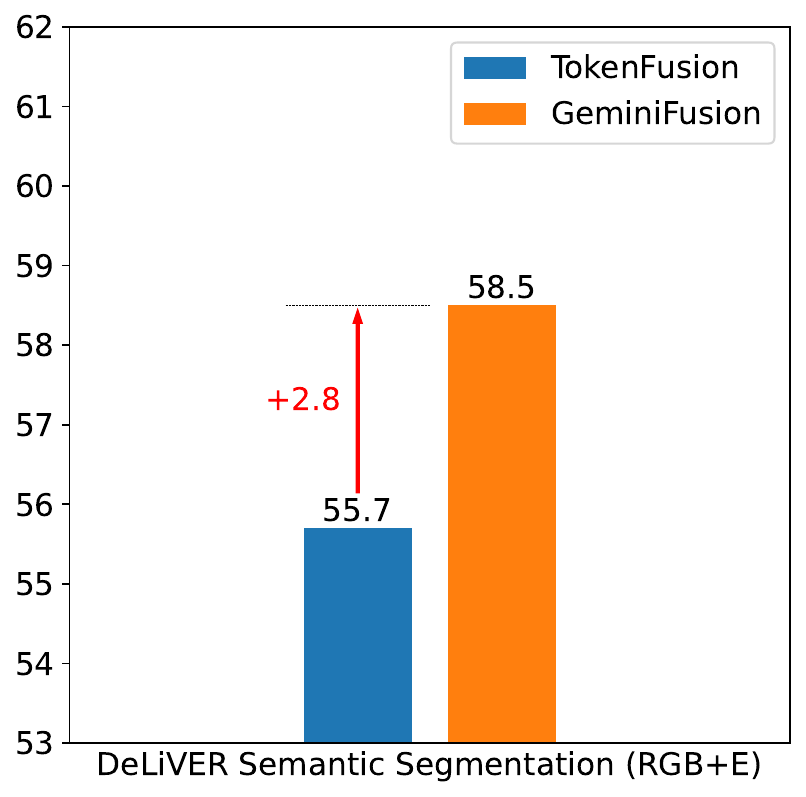}}
\subfloat{\includegraphics[height=34mm]{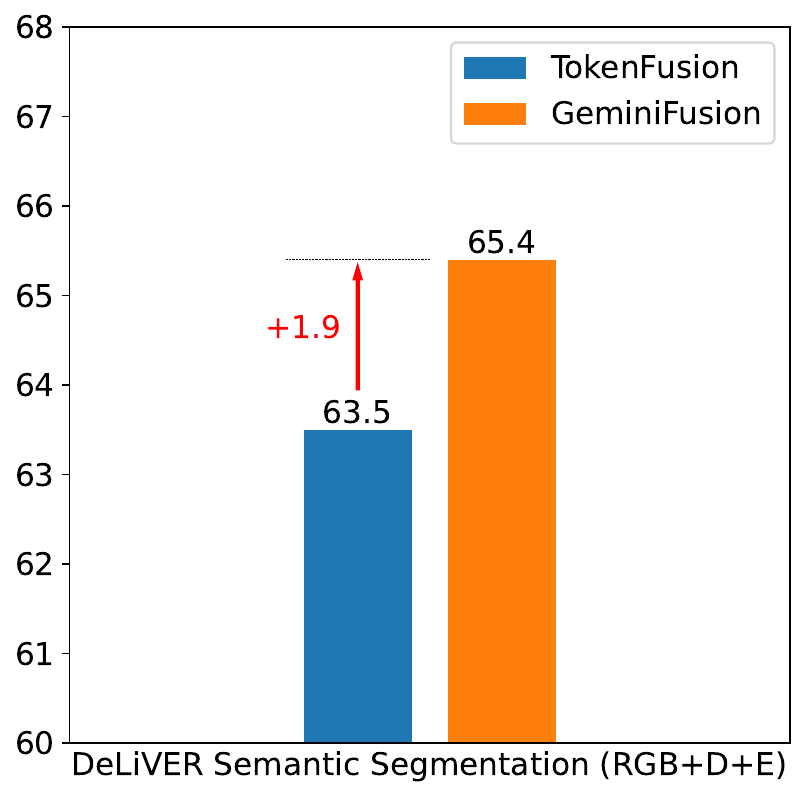}}
\subfloat{\includegraphics[height=34mm]{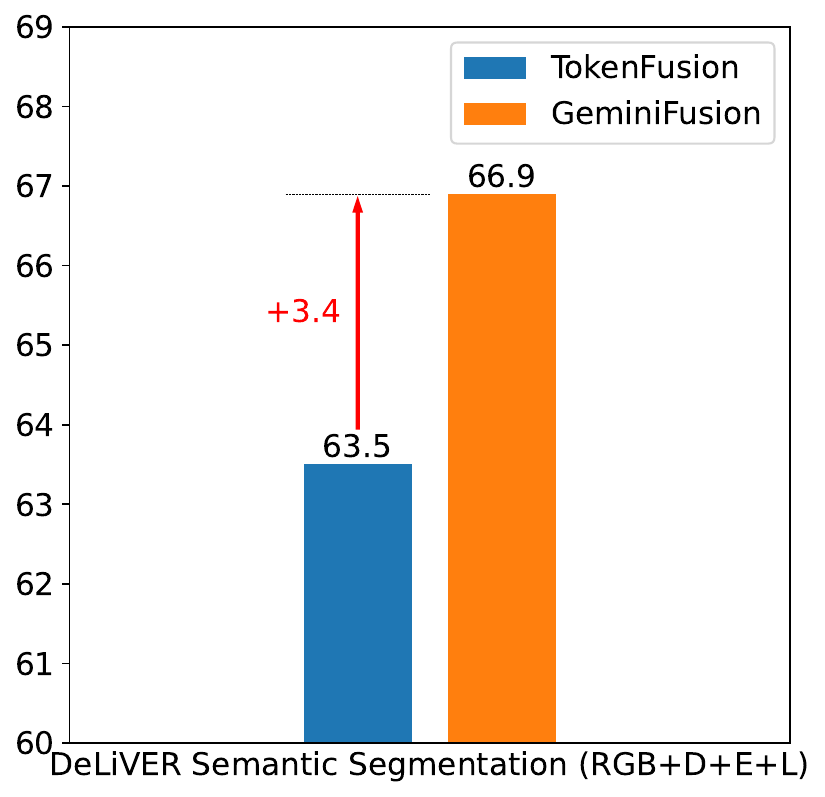}}
\vspace{-6pt}
\caption{\small{Improvements of our $\gemini$GeminiFusion across five multimodal semantic segmentation tasks. GeminiFusion achieves +$2.6\%$, +$1.3\%$, +$2.8\%$, +$1.9\%$, and +$3.4\%$ performance gains. All training epoch numbers are aligned. D: Depth, E: Event, L: LiDAR.}}
\label{fig:intro_results}
\end{figure*}

In light of the increasing availability of low-cost sensors, multimodal fusion which leverages data from various sources has emerged as a pivotal catalyst for advancing artificial intelligence-driven perception in vision~\cite{smith2005development,baltruvsaitis2018multimodal,cmt}. This approach has demonstrated remarkable potential, surpassing the unimodal paradigm across various downstream tasks, including autonomous driving~\cite{mfnet,bevformer}, semantic segmentation~\cite{ye2019cross,cao2021shapeconv}, video captioning~\cite{sun2019videobert,lu2019vilbert} and visual question answering~\cite{antol2015vqa,ben2017mutan}.

In the current literature, dominant paradigms for the multimodal fusion can be categorized into two ad-hoc schemes, \ie, interaction-based fusion~\cite{shvetsova2022everything,nagrani2021attention,cmx} and exchange-based fusion~\cite{wang2020deep,tokenfusion,zhu2023exchanging}.
In early interaction-based methods, a common practice involved directly concatenating tokens from different modalities~\cite{su2019vl}. This straightforward fusion approach neglects inter-modal interactions and sometimes leads to a poorer performance than single-modal counterparts~\cite{wang2020makes,tokenfusion}. While cross-attention mechanisms are introduced as a solution, the quadratic complexity of the full attention with an increasing number of input tokens challenges the feasibility of cross-modal models.
To tackle this issue, a simple strategy is to confine cross-modal interaction to later layers, often referred to as late-fusion~\cite{nagrani2021attention}. However, this method restricts the ability of the network's shallow layers to access valuable features from another modality, diminishing the original goal of facilitating mutual assistance between modalities and hindering overall model performance.

Exchange-based fusion provides a parameter-free solution~\cite{tokenfusion,wang2020deep} to the computational overhead by leveraging the inherent alignment of different modalities in vision tasks. For instance, world-space data like LiDAR and point clouds can be projected to pixels on the paired image plane. This method entails dynamically predicting the significance of each input token and subsequently replacing less crucial tokens from one modality with those from another.

\begin{figure*}[t]
\centering
\subfloat{\includegraphics[height=23mm]{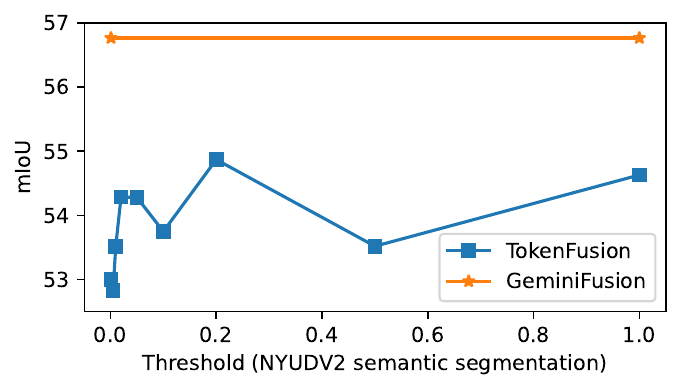}\label{fig:tf_th_nyu}}
\subfloat{\includegraphics[height=22.5mm]{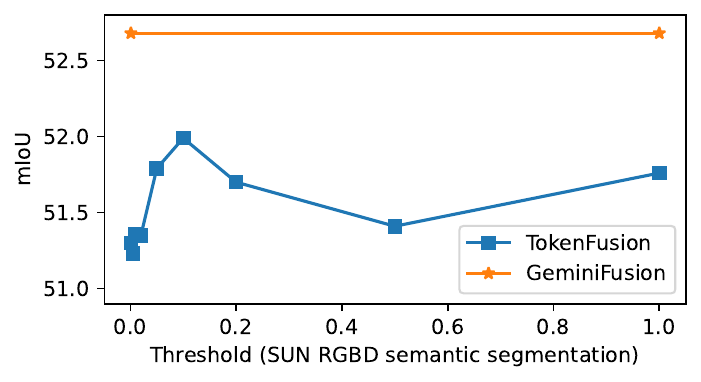}\label{fig:tf_th_sun}}
\subfloat{\includegraphics[height=22.5mm,width=42mm]{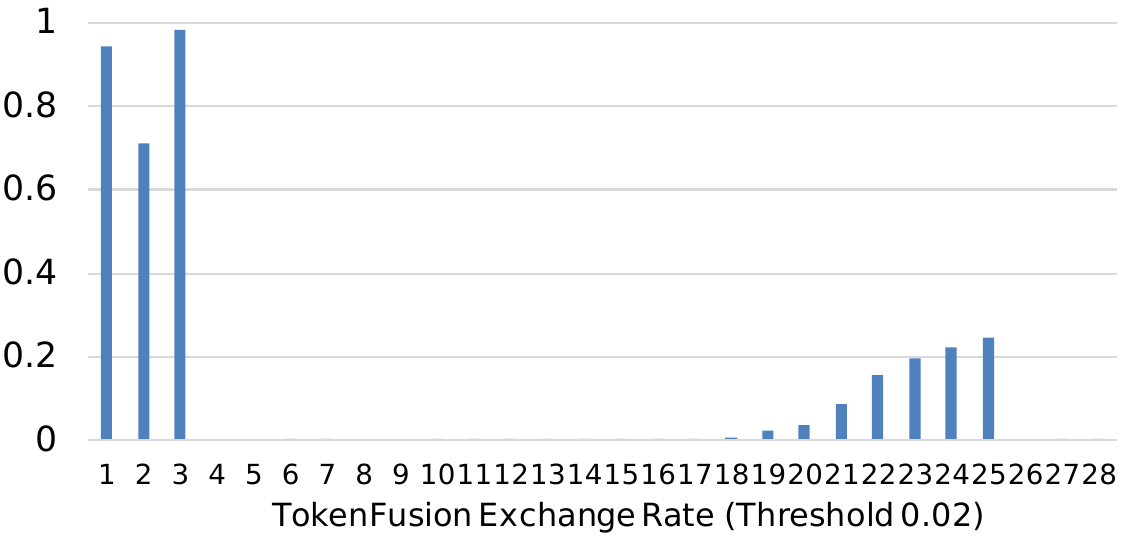}\label{fig:tf_ex_rate0.02}}
\subfloat{\includegraphics[height=22.7mm,width=42mm]{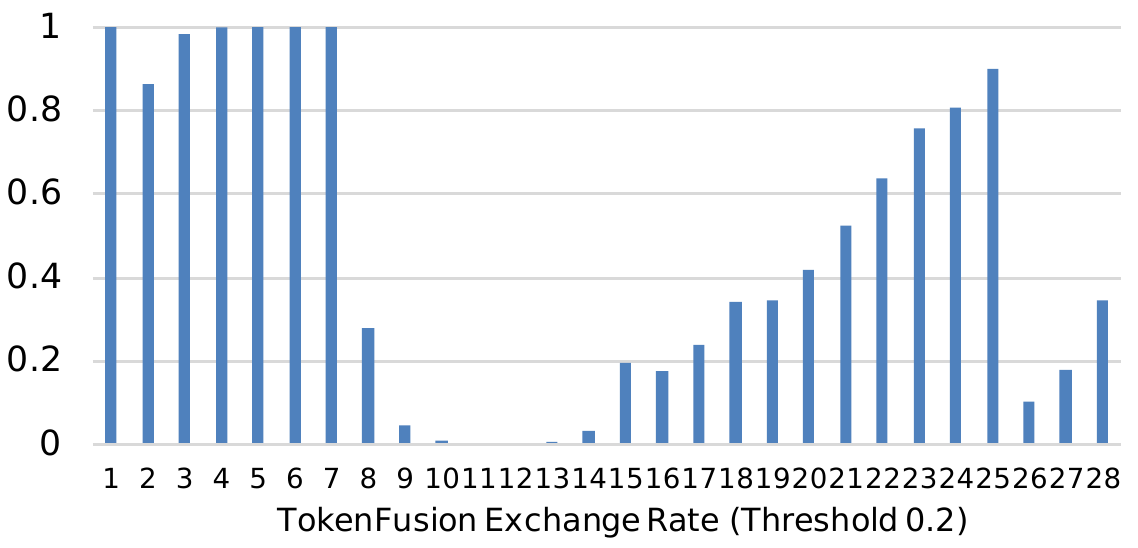}\label{fig:tf_ex_rate0.2}}
\vspace{-8pt}
\caption{\small{Impact of the threshold on the exchange-based TokenFusion. Exchanging all tokens almost invariably yields the best outcomes.}}
\label{fig:tokenfusion_threshold_curve}
\end{figure*}

Our investigation into the prune-then-substitute technique, as outlined in the TokenFusion~\cite{tokenfusion}, reveals that its effectiveness is not as consistent as expected. We observe that the network's shallow layers deem all tokens insignificant and indiscriminately substitute them with representations from an alternate modality. This behavior is in stark contrast to that of the deeper layers, which align more closely with our initial expectations by selectively swapping out representations of less pivotal tokens.
Moreover, our results suggest that a strategy of unconditionally exchanging all tokens almost invariably yields the best outcomes, as evidenced by the data presented in Figure~\ref{fig:tokenfusion_threshold_curve}. Upon further analysis, we believe that this phenomenon can be attributed to the intrinsic unique information carried by each token; any direct substitution results in an irrevocable loss of information. We also note instances of simultaneous information exchange at identical positions across modalities, underscoring the necessity for features from different modalities to be mutually retained and integrated.

We observe that the performance of the exchange-based fusion consistently underperforms the cross-attention based fusion, while the additional overhead introduced by the full attention poses a significant challenge.
To overcome this challenge and maintain the core information captured by the original unimodal learning, we introduce a pixel-wise multimodal fusion approach called GeminiFusion. Specifically, given two modalities, only the two matched tokens from corresponding modalities will participate in the fusion process. This fusion scheme has a minimal impact on the original unimodal representations, on account of the preservation of skip connections from the original inputs and the retention of self-consistent part during the fusion process. Meanwhile, the cross-modality part can significantly capture valuable multimodal information. The computational cost is minor since the pixel-wise attention is more compact compared to the full attention. Moreover, GeminiFusion demonstrates its superiority by allowing multimodal architectures to leverage parameters from unimodal pre-training, such as on the ImageNet dataset.

To verify the advantage of the proposed method, we consider extensive tasks including multimodal image-to-image translation, $3$D object detection and arbitrary-modal semantic segmentation, \ie, RGB, depth, events, and LiDAR, covering four multimodal benchmarks.

Our contributions in this paper include: (i) we empirically demonstrate that directly replacing features of one modality with those from another modality is sub-optimal. Simply exchanging all tokens every time achieves better results;
(ii) we propose an efficient method named GeminiFusion for multimodal feature fusion, leveraging the inherent high alignment of different modal inputs in vision tasks while preserving the original unimodal features;
(iii) extensive experiments on multimodal image-to-image translation, $3$D object detection tasks and arbitrary-modal segmentation consistently affirm the effectiveness of our proposed GeminiFusion.

\section{Related Work}

\begin{figure*}[t]
\centering
\includegraphics[width=0.95\linewidth]{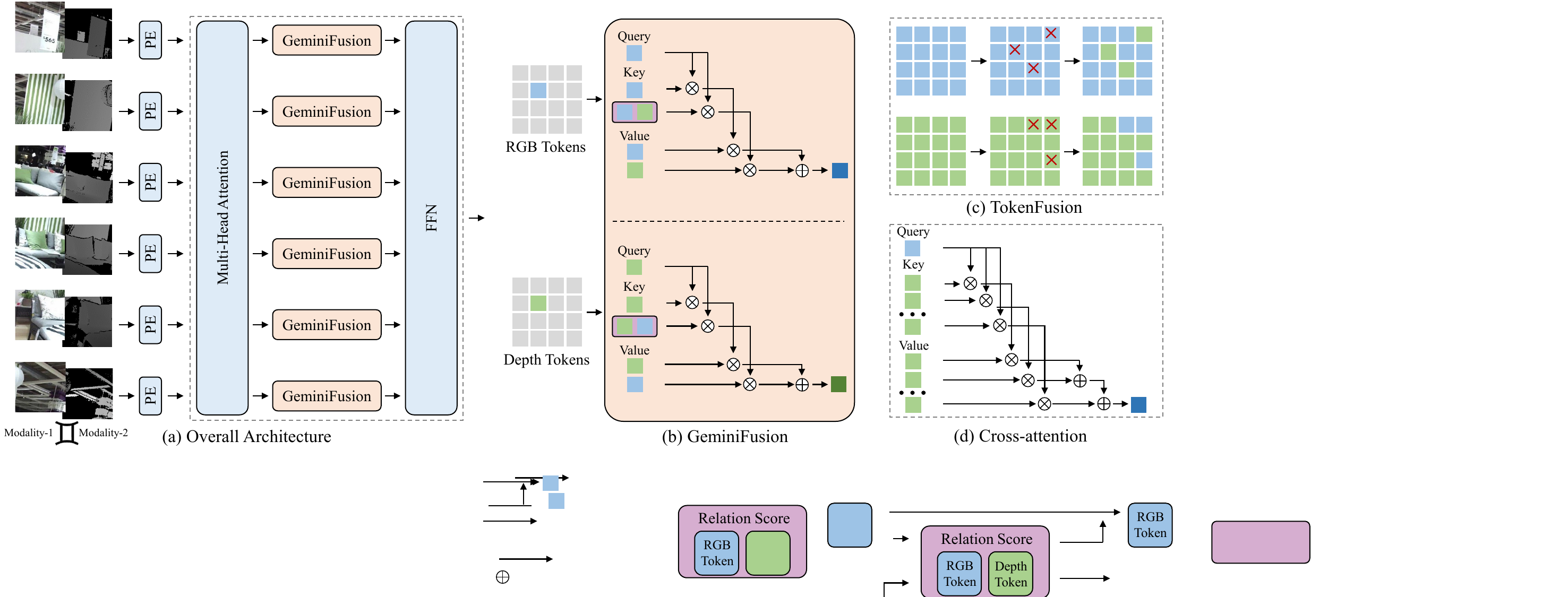}
\vspace{-2pt}
\caption{\small{(a) Overall architecture of GeminiFusion: our proposed GeminiFusion model is designed to be plug and play, allowing it to be seamlessly integrated into various vision backbones. (b) GeminiFusion module: performing pixel-wise fusion to enrich multimodal feature by utilizing aligned features from two modalities. (c) TokenFusion: swapping certain pixels between two features, but result in information loss. (d) Cross-attention: requires a significant amount of memory resources with quadratic complexity of input token.}}
\label{fig:framework}
\end{figure*}

The process of multimodal fusion involves leveraging diverse data sources to enhance associated details, surpassing the capabilities of their unimodal counterparts. Here, we delve into two prevailing fusion schemes and emphasizing their applicability in targeted multimodal vision tasks.

\noindent\textbf{Interaction-based multimodal fusion.} 
Early studies of interaction-based fusion~\cite{snoek2005early,atrey2010multimodal,bruni2014multimodal} categorizes the fusion strategy into three broad types: early (input-level), mid (feature-level) and late (decision-level) fusion. Early fusion methods~\cite{zhao2020single,zhang2021uncertainty} directly fuse the inputs from different modalities through a single-stream network, performed by averaging~\cite{hazirbas2017fusenet} or concatenating~\cite{zhang2018deep} along the input channels. However, the supervision signal is distant from the blended input, resulting in suboptimal results. Additionally, maintaining supervision for individual modalities is not feasible within this framework. Mid fusion~\cite{lin2017cascaded,chen2019selective,fu2020jl,ramachandram2017deep,de2017modulating} harnesses individual CNN or transformer encoders for each modality to capture intricacies in their respective features~\cite{xu2023fdvit,guo2022hire}. For example, MBT\cite{nagrani2021attention} subsequently amalgamated the encoded features through a dedicated fusion layer. RDFNet~\cite{park2017rdfnet} and CMX~\cite{cmx} employ multilayer fusion, aggregating features iteratively with additional convolutional blocks. EPIC-Fusion~\cite{kazakos2019epic} combines intermediate activations via summation in the joint training of multiple modality-specific networks. TransFuser~\cite{prakash2021multi} utilizes several transformer modules for the fusion of intermediate features between different modalities.
Late fusion~\cite{owens2018audio} aggregates the final decision through an ensemble of multiple outputs~\cite{pandeya2021deep,glodek2011multiple}, usually implemented using parallel networks.

\noindent\textbf{Exchange-based multimodal fusion.}
CEN~\cite{wang2020deep} introduces the parameter-free Channel Exchanging Network, which dynamically exchanges channels between sub-networks of different modalities. MLF-VO\cite{jiang2022self} extends this method to fuse color and inferred depth maps, incorporating a polarization regularizer to prevent the model from reaching a singular solution. MuSE~\cite{zhu2023exchanging} generalizes exchange-based methods from vision-vision fusion to text-vision fusion. TokenFusion~\cite{tokenfusion}, on the other hand, performs the exchange in the token dimension. It dynamically detects uninformative tokens and substitutes these tokens with features from other modalities. In this paper, we contend that the prune-then-substitute approach employed by TokenFusion consistently falls short in performance compared to the cross-attention-based interaction method. There is also a risk that all tokens undergo unnecessary exchange, resulting in irreversible information loss.

\noindent\textbf{Attention for multimodal fusion.}
Attention mechanisms, including self-/cross-attention~\cite{vaswani2017attention}, CBAM~\cite{cbam}, SENet~\cite{senet}, and ECA~\cite{wang2020eca} have demonstrated their success in various tasks. Several multimodal frameworks~\cite{bevformer,hori2017attention,wei2020multi} incorporate attention modules to fuse features from different modalities. For instance, ACNet~\cite{hu2019acnet} processes RGB and depth with two branches and employs the proposed Attention Complementary Module (ACM) to enable the fusion branch, exploiting more high-quality features from different channels. Different from the ACNet, we concentrate more on the aligned spatial location to explore an efficient fusion method. VST~\cite{liu2021visual} utilizes cross-attention to fuse features from two modalities by computing the self-attention between the queries from one modality and the keys and values from the other modality. TransFuser~\cite{prakash2021multi} and TriTransNet~\cite{liu2021tritransnet} concatenate two modal features and use self-attention to mix information. Additionally, works like~\cite{zhao2021rgb,wang2022learning} employ the SE module to blend information. In contrast to previous quadratic complexity cross-attention, our pixel-wise attention has linear complexity with respect to the number of input tokens. This feature enables our fusion method to maintain a nearly as compact multimodal architecture as a unimodal network.

\noindent\textbf{Multimodal semantic segmentation.}
Many segmentation methods excel in standard RGB-based benchmarks, providing per-pixel category predictions in a given scene. However, they often face challenges in real-world scenarios with rich $3$D geometric information. To overcome this limitation, researchers have sought to enhance scene understanding by incorporating multimodal sensing, including depth~\cite{nyu, gupta2014learning}, thermal~\cite{mfnet,sun2019rtfnet}, polarimetric optical cues~\cite{kalra2020deep}, event-driven priors~\cite{zhang2021exploring}, and LiDAR~\cite{zhuang2021perception,nuscenes}. Previous works have primarily focused on the RGB-depth setting, which may not generalize well across different sensing data~\cite{cmx}. In this study, we explore a unified approach capable of generalizing effectively to diverse multimodal combinations for semantic segmentation.

\begin{figure}[t]
\centering
\includegraphics[width=1\linewidth]{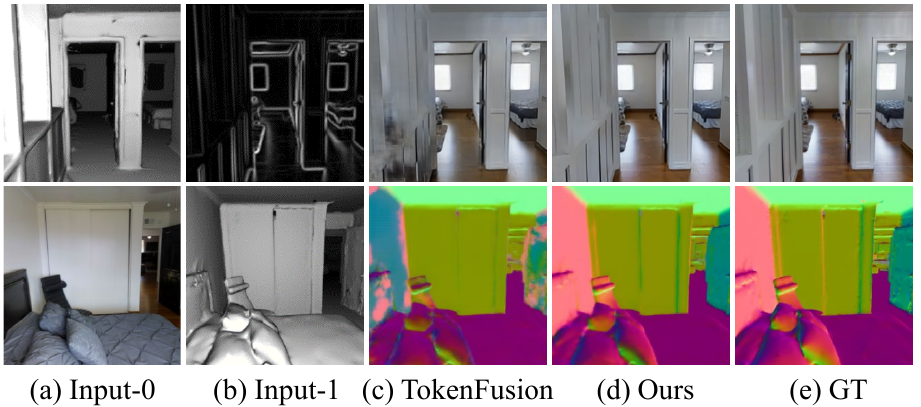}
\vspace{-18pt}
\caption{\small{Image-to-image translation results on the validation split of Taskonomy. Best view in color and zoom in.}}
\label{fig:vis_img2img}
\end{figure}

\section{Our Method}
We first revist the recently proposed TokenFusion~\cite{tokenfusion} method in Section~\ref{sec:tokenfusion}. Subsequently, Section~\ref{sec:cross-attn} details the commonly utilized cross-attention mechanism. Our pixel-wise GeminiFusion module is introduced in Section~\ref{sec:Geminifusion}, and the comprehensive architecture is presented in Section~\ref{sec:arch}.

\subsection{Fusion via exchange}
\label{sec:tokenfusion}
Based on the motivation that there are always uninformative tokens or channels in single-modal transformers, exchange based methods such as TokenFusion~\cite{tokenfusion} and CEN~\cite{wang2020deep} are designed to dynamically detect and substitute these useless tokens or channels with features from other modalities. Specifically, at the core of its functionality, TokenFusion~\cite{tokenfusion} prunes tokens in each modality and replaces them with corresponding tokens from other modalities that have been projected and aggregated to match. This exchange is guided by a score predictor integrated within each block of the network, which computes masks that share the dimensions of the multimodal inputs. These masks, through a comparison against a predefined threshold, facilitate the selection of tokens to be substituted. Specifically, if there are only two modalities as input, \ie, $\mathbf{X}^{\rm1}$ and $\mathbf{X}^{\rm2}$, the token exchange process can be formulated as:
\begin{equation}
\small
\begin{aligned}
\mathbf{X}^1_{\rm[i]}=\mathbf{X}^1_{\rm[i]}\odot\mathbb{I}_{s(\mathbf{X}^1_{\rm[i]})\ge\theta} + \mathbf{X}^2_{\rm[i]}\odot\mathbb{I}_{s(\mathbf{X}^1_{\rm[i]})<\theta}, \\
\mathbf{X}^2_{\rm[i]}=\mathbf{X}^2_{\rm[i]}\odot\mathbb{I}_{s(\mathbf{X}^2_{\rm[i]})\ge\theta} + \mathbf{X}^1_{\rm[i]}\odot\mathbb{I}_{s(\mathbf{X}^2_{\rm[i]})<\theta}.
\end{aligned}
\end{equation}

where $\mathbf{X}^1_{\rm[i]}$ indicates the $i$-th token of input $\mathbf{X}^{\rm1}$, $\mathbb{I}$ is an indicator asserting the subscript condition, therefore it outputs a mask tensor $\in\{0,1\}^N$, the parameter $\theta$ is a small threshold set to 0.02, and the operator $\odot$ resents the element-wise multiplication.

The supervision of the mask generation process is enforced through an $L$-1 norm constraint. However, this approach introduces an element of stochasticity. The model does not inherently prioritize the informational importance of tokens when generating the masks. We contend that the connection between the masks and the tokens' intrinsic information content is not well-regulated, which may lead to randomness in the exchange process. As demonstrated in Figure~\ref{fig:tf_ex_rate0.02} and Figure~\ref{fig:tf_ex_rate0.2}, altering the threshold does not prevent the tokens in the initial layers from being entirely exchanged. This suggests that TokenFusion does not operate as initially hoped, where tokens with negligible information are replaced by those from other modalities. Furthermore, as illustrated in Figure~\ref{fig:tf_th_nyu} and Figure~\ref{fig:tf_th_sun}, setting the threshold to 1, thereby allowing all tokens always to be exchanged, yields better results. This indicates that the exchange-based method of TokenFusion is not only unstable but also prone to the loss of critical information. Hence, it may be less effective than a strategy involving the complete exchange of information.

\begin{table}[t]
\newcommand{\blue}[1]{\textcolor{blue}{#1}}
\definecolor{deepgreen}{rgb}{0.07, 0.53, 0.03}
\centering
\setlength{\tabcolsep}{2pt}
\renewcommand{\arraystretch}{1.2}
\footnotesize
\caption{\small{Comparison with TokenFusion on the NYUDv2, SUN RGB-D and DeLiVER datasets for multimodal semantic segmentation task. Evaluation metrics include pixel accuracy (\%),
mean accuracy (\%), and mean IoU (\%). Only mIoU is reported on the DeLiVER dataset following CMNeXt~\cite{deliver}. $^\dag$ marks the methods are reproduced by ourselves. All training epochs are aligned. D: Depth, E: Event, L: LiDAR.}
\vspace{-4pt}
}
\label{tab:tf_ge_seg}
\resizebox{1.0\linewidth}{!}
{
\begin{tabular}{l|c|c|ccc}
\shline
Method    & Backbone & Inputs &  Pixel Acc. &  mAcc. & mIoU \\
\shline
\multicolumn{6}{l}{\emph{Results on the NYUDv2 dataset}}  \\\hline
TokenFusion  & MiT-B$3$   & RGB+D     & $79.0$ & $66.9$ & $54.2$ \\
GeminiFusion    & MiT-B$3$   & RGB+D     & $\bf{79.9}^{\textrm{+0.9}}$ & $\bf{69.9}^{\textrm{+3.0}}$ & $\bf{56.8}^{\textrm{+2.6}}$ \\
\shline
TokenFusion{$^\dag$}  & MiT-B$5$   & RGB+D     & $79.1$ & $67.5$ & $55.1$ \\
GeminiFusion    & MiT-B$5$   & RGB+D     & $\bf{80.3}^{\textrm{+$1.2$}}$ & $\bf{70.4}^{\textrm{+$2.9$}}$ & $\bf{57.7}^{\textrm{+$2.6$}}$ \\
\shline
\multicolumn{6}{l}{\emph{Results on the SUN RGB-D dataset}}  \\\hline
TokenFusion{$^\dag$}  & MiT-B$3$   & RGB+D     & $82.8$ & $63.6$ & $51.4$ \\
GeminiFusion    & MiT-B$3$   & RGB+D     & $\bf{83.3}^{\textrm{+0.5}}$ & $\bf{64.6}^{\textrm{+1.0}}$ & $\bf{52.7}^{\textrm{+1.3}}$ \\
\shline
TokenFusion{$^\dag$}  & MiT-B$5$   & RGB+D     & $83.1$ & $63.9$ & $51.8$ \\
GeminiFusion    & MiT-B$5$   & RGB+D     & $\bf{83.8}^{\textrm{+0.7}}$ & $\bf{65.3}^{\textrm{+1.4}}$ & $\bf{53.3}^{\textrm{+1.5}}$ \\
\shline
\multicolumn{6}{l}{\emph{Results on the DeLiVER dataset}}  \\\hline
TokenFusion{$^\dag$}  & MiT-B$2$   & RGB+D     & - & - & $63.7$ \\
GeminiFusion    & MiT-B$2$   & RGB+D     & - & - & $\bf{66.4}^{\textrm{+2.7}}$ \\
\shline
TokenFusion{$^\dag$}  & MiT-B$2$   & RGB+E     & - & - & $55.7$ \\
GeminiFusion    & MiT-B$2$   & RGB+E     & - & - & $\bf{58.5}^{\textrm{+2.8}}$ \\
\shline
TokenFusion{$^\dag$}  & MiT-B$2$   & RGB+L     & - & - & $55.5$ \\
GeminiFusion    & MiT-B$2$   & RGB+L     & - & - & $\bf{58.6}^{\textrm{+3.1}}$ \\
\shline
TokenFusion{$^\dag$}  & MiT-B$2$   & RGB+D+E+L     & - & - & $63.5$ \\
GeminiFusion    & MiT-B$2$   & RGB+D+E+L     & - & - & $\bf{66.9}^{\textrm{+3.4}}$ \\
\shline
\end{tabular}
}
\end{table}

\begin{table}
\begin{minipage}[t]{1\linewidth}
\small
\centering
\setlength{\tabcolsep}{8pt}
\renewcommand{\arraystretch}{1.0}
\caption{\small{Comparison on the Taskonomy dataset for the multimodal image-to-image translation task. Evaluation metrics are FID/KID (×10$^{-2}$) for the RGB predictions and MAE (×$10^{−1}$)/MSE (×$10^{−1}$) for other predictions. Lower values indicate better performance for all the metrics. All training epoch numbers are aligned.}}
\label{tab:tf_ge_img2img}
\vspace{-4pt}
\resizebox{1.0\linewidth}{!}
{
    \begin{tabular}{l|ccccc}
        \shline
        Method &  \makecell[c]{Shade+Texture \\$\rightarrow$ RGB} & \makecell[c]{Depth+Normal \\$\rightarrow$ RGB} & \makecell[c]{RGB+Shade \\$\rightarrow$ Normal}  & \makecell[c]{RGB+Edge \\$\rightarrow$ Depth}               \\
        \shline
        TokenFusion & $47.31 / 0.94$     & $103.87/4.24$  & $0.67 / 1.75$ & $0.22 / 0.55$  \\
        \shline
        
        GeminiFusion   & $\bf{41.32}^{\textrm{-5.99}} / \bf{0.81}^{\textrm{-0.13}}$     & $\bf{96.98}^{\textrm{-6.89}} / \bf{3.71}^{\textrm{-0.53}}$  & $\bf{0.65}^{\textrm{-0.02}} / \bf{1.69}^{\textrm{-0.06}}$ & $\bf{0.20}^{\textrm{-0.02}} / \bf{0.49}^{\textrm{-0.06}}$  \\
        \shline
    \end{tabular}
            }
\vspace{3mm}
\end{minipage}

\begin{minipage}[t]{1\linewidth}
\small
\centering
\setlength{\tabcolsep}{8pt}
\renewcommand{\arraystretch}{1.0}
        \caption{\small{
            Comparison with MVX-Net on the 3D object detection task against vehicle targets. The dataset is the validation set of the KITTI 3D object detection dataset. All training epoch numbers are aligned. The IoU threshold is 0.7.
        }}
        \label{tab:3d_det}
        \resizebox{1.0\linewidth}{!}
{
            \begin{tabular}{c|c|c|c|c|c|c|c}
                \shline
                \multirow{2}{*}{Method}    & \multirow{2}{*}{Param(M)} & \multicolumn{3}{c|}{3D $AP_{R11}$} & \multicolumn{3}{c}{3D $AP_{R40}$}                                                             \\\cline{3-8}
                                       &                         & Easy                                 & Medium                                & Hard           & Easy                                 & Medium                                & Hard           \\\shline
                MVX-Net                & $33.8$                  & $87.49$                            & $77.04$                           & $74.54$      & $88.41$      & $\bf{78.77}$ & $74.27$      \\
                MVX-Net + GeminiFusion & $34.8$                  & $\bf{88.49}$                       & $\bf{77.36}$                      & $\bf{74.61}$ & $\bf{89.43}$ & $78.76$      & $\bf{74.46}$ \\
                \shline
            \end{tabular}
            }
\end{minipage}
\end{table}

\subsection{Fusion via cross-attention}
\label{sec:cross-attn}
We commence with an exploration of a prevalent cross-attention-based fusion architecture~\cite{bevformer,detr}, which is typified by the utilization of a canonical attention scheme to process inputs derived from multiple modalities. As illustrated in Figure~\ref{fig:framework}d, consider the scenario where we have procured a set of $N$ patches from two modalities, denoted as $\mathbf{X}^1, \mathbf{X}^{\rm 2}\in\mathbb{R}^{N\times d}$, the corresponding output $\mathbf{Y}^{\rm 1}, \mathbf{Y}^{\rm 2}\in\mathbb{R}^{N\times d}$ augmented by multimodal information can be generated by:
\begin{equation}
\small    
\begin{aligned}
&\mathbf{Y}^{\rm 1} = \rm{Attention}(\mathbf{X}^{\rm 1}W^Q,\mathbf{X}^{\rm 2}W^K,\mathbf{X}^{\rm 2}W^V) +  \mathbf{X}^{1} \\
&\mathbf{Y}^{\rm 2} = \rm{Attention}(\mathbf{X}^{\rm 2}W^Q,\mathbf{X}^{\rm 1}W^K,\mathbf{X}^{\rm 1}W^V) +  \mathbf{X}^{2} \\
&\rm{Attention}(Q,K,V)=\rm{Softmax}(QK^T/\sqrt{d})V
\end{aligned}
\label{equ:attn}
\end{equation}
The computational complexity of above operation is $\mathcal{O}(N^2\cdot c)$, where $N$ is the number of tokens of both modalities. Given that $N$ can be exceptionally large, the computational demand of the model is significantly increased. For instance, CMNeXt~\cite{deliver} partitions each modality input into $16,384$ patches. This partitioning leads to a computational requirement of over $17$G FLOPs for just one instance of cross-attention, a figure that is prohibitive for practical model deployment.

\subsection{$\gemini$GeminiFusion: pixel-wise fusion module}
\label{sec:Geminifusion}
To harness the benefits of modality fusion through cross-attention mechanism while circumventing the computational intensity that it entails, we introduce an innovative pixel-wise fusion module, termed the GeminiFusion module.

\begin{figure}[t]
\centering
\subfloat{\includegraphics[height=21mm]{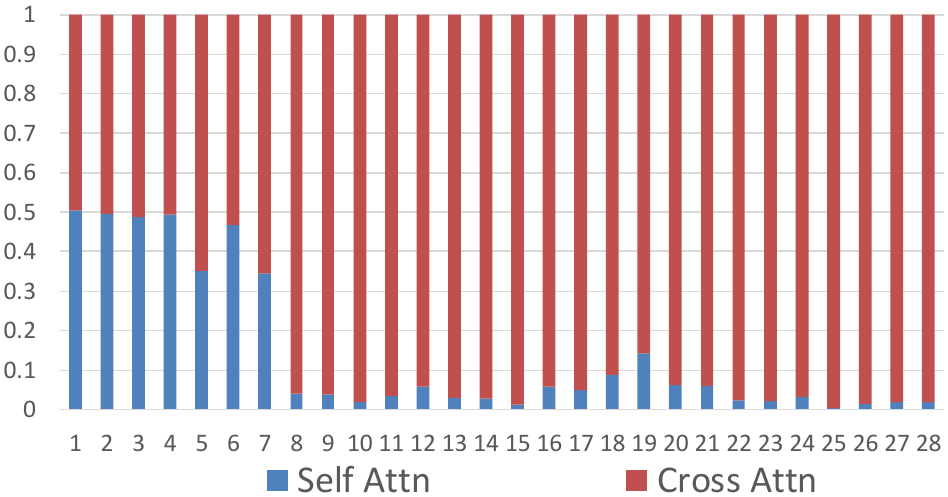}}
\subfloat{\includegraphics[height=21mm]{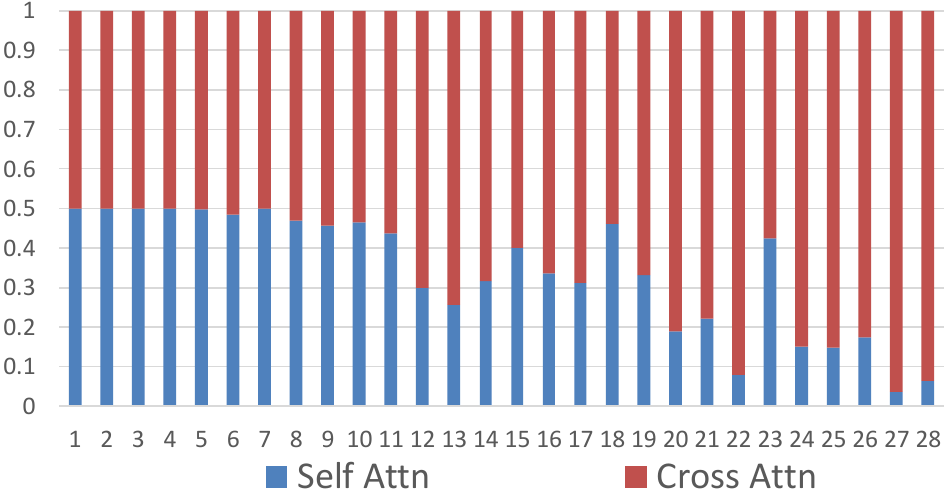}}
\vspace{-6pt}
\caption{\small{Comparison of attention scores obtained from self-attention (intra-modality) and cross-attention (inter-modality). Left: with noise. Right: without noise.}}
\label{fig:gemini_attn_score}
\end{figure}

Drawing inspiration from TokenFusion~\cite{tokenfusion}, we posit that not all patches contribute equally to the fusion process. Less salient patches could be efficiently substituted by their spatial counterparts from the alternate modality, implying that exhaustive interaction among all patches may not be obligatory. This insight leads us to the hypothesis that the crux of inter-modality information exchange lies in the patches sharing identical spatial coordinates, as these locations are where information exchange is most pertinent and significant. Leveraging this insight, the GeminiFusion module is engineered to prioritize interactions between spatially co-located patches from different modalities, thus refining the cross-attention mechanism:
\begin{equation}
\footnotesize
\begin{aligned}
&\mathbf{Y}^1_{\rm[i]}=\rm{Attention}(\mathbf{X}^1_{\rm[i]}W^Q,\mathbf{X}^2_{\rm[i]}W^K,\mathbf{X}^2_{\rm[i]}W^V) + \mathbf{X}^1_{\rm[i]}, \\
&\mathbf{Y}^2_{\rm[i]}=\rm{Attention}(\mathbf{X}^2_{\rm[i]}W^Q,\mathbf{X}^1_{\rm[i]}W^K,\mathbf{X}^1_{\rm[i]}W^V) + \mathbf{X}^2_{\rm[i]}.
\end{aligned}
\end{equation}
where $\rm i$ is in the range of $d$. The targeted interaction strategy of GeminiFusion module not only focuses computational effort on the most critical information exchanges but also significantly slashes the computational load. This efficiency is quantified by a reduction in computational complexity to $\mathcal{O}(N\cdot c^2)$. Compared with the cross-attention, the FLOPs plummet from $17$G to merely $0.14$G. This staggering reduction of $99.2\%$ in computational demand marks a transformative improvement, rendering the module exceedingly efficient for deployment in environments where computational resources are at a premium or where real-time performance is necessary.

However, two main challenges arise here: \textbf{(i) Incongruity outcomes from the attention score.} In the TokenFusion~\cite{tokenfusion} framework, the exchange of less informative patches with those from a different modality has been shown to enhance model performance. Conversely, within the attention module, a tendency arises where one modality disproportionately learns from patches of another modality that are more self-similar, as they yield higher attention scores. This proclivity is antithetical to our intended model behavior, which seeks to benefit from the integration of dissimilar and potentially more informative patch characteristics. \textbf{(ii) Softmax function limitation in per-pixel attention mechanism.} The current attention formulation operates on a per-pixel basis, resulting in an attention map of dimension $1 \times 1$. The application of the softmax function in this context is rendered ineffective as it invariably returns a value of one, nullifying the intended differentiation of the attention mechanism. This outcome undermines the capacity of the model to assign varying levels of attention across modalities.

To address the aforementioned issues, we propose two enhancements. Firstly, we introduce a lightweight \emph{relation discriminator} to evaluate the disparity between modalities. Our findings indicate that a synergistic combination of a $1 \times 1$ convolution followed by a softmax function suffices. The associated experiments are detailed in Table~\ref{tab:ab_relation_discriminator}. Specifically, patches from the two modalities are concatenated and fed into the relation discriminator, which subsequently assigns a relation score ranging from $0$ to $1$. This relation score is utilized to modulate the original key, effectively substituting the standard key in Eq.~\ref{equ:attn}:
\begin{equation}
\footnotesize
\begin{aligned}
    &\mathbf{Y}^{\rm1}_{\rm[i]}=\rm{Attention}(Q,K,V)+\mathbf{X}^{\rm1}_{\rm[i]} \\
    &\rm Q=\mathbf{X}^{\rm1}_{\rm[i]}W^Q,\;\rm K=\mathbf{X}^{1}_{\rm[i]}\phi(\mathbf{X}^{1}_{\rm[i]},\mathbf{X}^2_{\rm[i]})W^K,\;\rm V=\mathbf{X}^2_{\rm[i]}W^V
\end{aligned}
\label{equ:relation_score}
\end{equation}
where $\phi(\cdot)$ indicates our relation discriminator module. The formula for $\rm Y^2_{\rm[i]}$ is obtained in the same way. To prevent the second issues associated with single-item focus without adding redundant information, we add the pixel-wise self-attention into the Eq.~\ref{equ:relation_score}:
\begin{equation}
\small
\begin{aligned}
    &\rm K=[\mathbf{X}^1_{\rm[i]}W^K,\,\,\mathbf{X}^1_{\rm[i]}\phi(\mathbf{X}^1_{\rm[i]},\mathbf{X}^2_{\rm[i]})W^K], \\
    &\rm V=[\mathbf{X}^1_{\rm[i]}W^V,\,\,\mathbf{X}^2_{\rm[i]}W^V].
\end{aligned}
\label{equ:relation_score_adj1}
\end{equation}
The formula for $\rm Y^2_{\rm[i]}$ is obtained in the same way. In the self-attention mechanism described by Equation~\ref{equ:relation_score_adj1}, both the query and key are derived from identical modal inputs, leading to an inherent bias towards the self-referential component of the attention score. This can diminish the efficacy of learning cross-modal representations. To address this issue, we augment the self-attention with \emph{layer-adaptive noise}. This approach involves the injection of a minimal amount of noise at the layer level, subtly enhancing the feature representation without burdening the model with extraneous information. To encapsulate this process for input tensors $\mathbf{X}^{\rm 1},\mathbf{X}^{\rm 2}\in\mathbb{R}^{N\times d}$ at Layer $L$, the resultant output tensors $\mathbf{Y}^{\rm 1},\mathbf{Y}^{\rm 2}\in\mathbb{R}^{N\times d}$ within our GeminiFusion module can be mathematically represented as follows:
\begin{equation}
\footnotesize
\begin{aligned}
    \mathbf{Y}^1_{\rm[i]}&=\rm{Attention}(Q^1,K^1,V^1) + \mathbf{X}^1_{\rm[i]}, \\
    \rm Q^1&=\mathbf{X}_{\rm 1[i]}W^Q, \\
    \rm K^1&=[(\rm{Noise}^{K}_{L}+\mathbf{X}^1_{\rm[i]})W^K,\,\,\mathbf{X}^1_{\rm[i]}\phi(\mathbf{X}^1_{\rm[i]},\mathbf{X}^2_{\rm[i]})W^K], \\
    \rm V^1&=[(\rm{Noise}^{V}_{L}+\mathbf{X}^1_{\rm[i]})W^V,\,\,\mathbf{X}^2_{\rm[i]}W^V] \\
    \mathbf{Y}^2_{\rm[i]}&=\rm{Attention}(Q^2,K^2,V^2)+ \mathbf{X}^2_{\rm[i]}, \\
    Q^2&=\mathbf{X}^2_{\rm[i]}W^Q, \\
    \rm K^2&=[(\rm{Noise}^{K}_{L}+\mathbf{X}^2_{\rm[i]})W^K,\,\,\mathbf{X}^2_{\rm[i]}\phi(\mathbf{X}^2_{\rm[i]},\mathbf{X}^1_{\rm[i]})W^K], \\
    \rm V^2&=[(\rm{Noise}^{V}_{L}+\mathbf{X}^2_{\rm[i]})W^V,\,\,\mathbf{X}^1_{\rm[i]}W^V].
\end{aligned}
\label{equ:final_attn_represent}
\end{equation}

We have conducted an ablation study on noise selection, detailed in Table~\ref{tab:ab_noise}. Our findings indicate that the optimal noise implementation involves a learnable parameter added to the key, with this parameter being unique to each layer. This layer-specific noise facilitates a dynamic balance between self-attention and cross-modal attention and ensures the appropriate functioning of the softmax operation. Figure~\ref{fig:gemini_attn_score} illustrates the variation in attention scores across increasing layer depths.

\begin{figure}[t]
\centering
\includegraphics[width=0.95\linewidth]{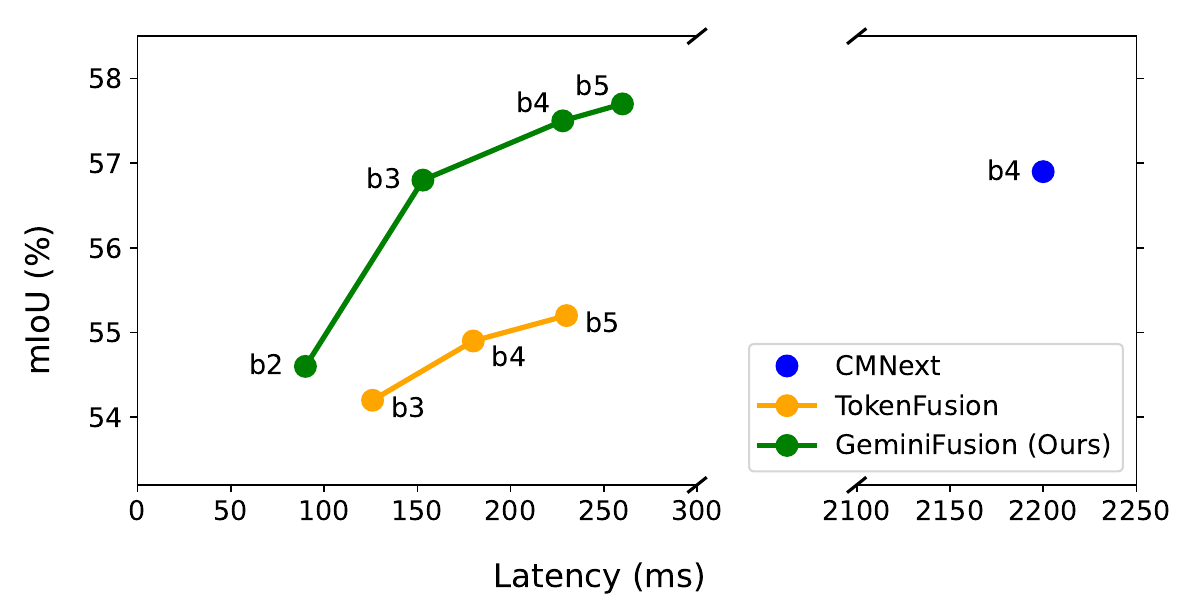}
\vspace{-4pt}
\caption{\small{Performance \vs latency on the NYUDv2 dataset. GeminiFusion achieves the better trade-off compared with others. Latency is measured by averaging all validation samples of the NYUDv2 dataset. Multi-scale flip test strategy is used in CMNext as described in ~\cite{deliver}.}}
\vspace{-8pt}
\label{fig:latency_miou}
\end{figure}

\subsection{Overall architecture}
\label{sec:arch}
Our GeminiFusion model adopts an encoder-decoder architecture, with the encoder featuring a four-stage structure akin to the widely recognized SegFormer~\cite{segformer} for the extraction of hierarchical features. For conciseness, Figure~\ref{fig:framework} illustrates only the initial stage out of the four.

The primary focus lies in multimodal fusion based on visual data, encompassing modalities such as RGB, depth, event, and LiDAR. These modalities are inherently homogeneous, as they represent different visual perspectives of the same subject, and can be readily converted into image-like formats~\cite{zhuang2021pmf,deliver}. Within our framework, all modalities utilize shared parameters with the exception of the Layer Normalization (LN) layers, facilitating a uniform processing approach. More specifically, the RGB image $\boldsymbol{I}_{RGB}\in\mathcal{R}^{3\times H\times W}$, along with the other $M-1$ modalities $\boldsymbol{I}_{depth},\cdots,\boldsymbol{I}_{LiDAR}\in\mathcal{R}^{3\times H\times W}$, undergoes sequential refinement through Multi-Head Self-Attention (MHSA) and Feed-Forward Network (FFN) blocks. These modalities are then adeptly integrated to harness intra-modal information via our proposed GeminiFusion module.

\begin{table}[t]
\begin{minipage}[t]{1\linewidth}
\small
\centering
\setlength{\tabcolsep}{8pt}
\renewcommand{\arraystretch}{1.0}
        \caption{\small{
            Comparison of multimodal semantic segmentation results on NYUDv2 and SUN RGBD datasets with Swin~\cite{swin} and MiT-B3/B5~\cite{segformer} as encoder models. All training epochs are aligned. Swin-Tiny-1k and Swin-Large-22k are pre-trained on the ImageNet-1K and ImageNet-22k, respectively.
        }}
        \label{tab:swin_seg}
        \resizebox{1.0\linewidth}{!}
{
            \begin{tabular}{l|l|c|c|c}
                \shline
                    Method                            & Encoder            & Param(M) & \makecell[c]{NYUDv2               \\ mIoU}            & \makecell[c]{SUNRGBD               \\ mIoU}                        \\
                \shline
                \multirow{4}{*}{GeminiFusion} & MiT-B3         & $75.8$ & $56.8$              & $52.7$      \\
                                              & MiT-B5         & $137.2$  & $57.7$              & $53.3$      \\ \cline{2-5}
                                              & Swin-Tiny-1k   & $52.0$ & $52.2$              & $50.2$      \\
                                              & Swin-Large-22k & $369.2$  & $\bf{60.2}$         & $\bf{54.6}$ \\
                \shline
            \end{tabular}
            }
\end{minipage}
\end{table}

\begin{table}[t]
\newcommand{\blue}[1]{\textcolor{blue}{#1}}
\definecolor{deepgreen}{rgb}{0.07, 0.53, 0.03}
\centering
\small
\setlength{\tabcolsep}{4pt}
\renewcommand{\arraystretch}{1.05}
\caption{\small{Comparison results with state-of-the-art methods on the NYUDv2, SUN RGB-D and DeLiVER datasets for the multimodal semantic segmentation task. Additional strategies indicate that the method uses strategies other than ImageNet classification pre-training. For the DeLiVER dataset, we follow CMNeXt to use MiT-B2 as backbone for fair comparison. Therefore ``MiT-B5 (MiT-B2)" indicates that we use MiT-B5 for NUYDv2 and SUN RGB-D, while MiT-B2 for DeLiVER. $^*$ indicates that we use the SUN RGBD trained model as pre-training on NYUDv2 dataset. $^\dag$ indicates that the results are reproduced by ourselves.}
\vspace{-5pt}
}
\label{tab:sota_results}
\resizebox{1.0\linewidth}{!}
{
\begin{tabular}{l|l|c|c|c|c}
                \shline
                Method                   & Backbone                & \makecell[c]{Additional \\ Strategies}  & \makecell[c]{NYUDv2                             \\ mIoU} & \makecell[c]{SUN RGBD               \\ mIoU} & \makecell[c]{DeLiVER               \\ mIoU} \\
                \shline
                PSD                  & ResNet$50$          & \xmark & $51.0$              & $50.6$      & -           \\
                FSFNet               & ResNet-$101$        & \xmark & $52.0$              & $50.6$      & -           \\
                TokenFusion{$^\dag$} & MiT-B$5$ (MiT-B$2$) & \xmark & $55.1$              & $51.8$      & $63.5$      \\
                SMMCL                & SegNeXt-B           & \xmark & $55.8$              & -           & -           \\
                MultiMAE             & ViT-Base            & \cmark & $56.0$              & -                & -         \\
                OMNIVORE             & Swin-Large          & \cmark & $56.8$              & -              & -           \\
                CMNeXt               & MiT-B$4$ (MiT-B$2$) & \xmark & $56.9$              & $50.4$      & $66.3$      \\
                CMX                  & MiT-B$5$            & \xmark & $56.9$              & $52.4$      & $62.7$      \\
                DFormer              & DFormer-L           & \cmark & $57.2$              & $52.5$      & -           \\
                PolyMaX              & ConvNeXt-L          & \cmark & $58.1$              & -              & -           \\
                SwinMTL              & SwinV2-Base-MiM     & \cmark & $58.1$              & -              & -           \\
                EMSANet              & EMSANet-R34-NBt1D   & \cmark & $59.0$              & $50.9$        & -            \\
                DPLNet               & MiT-B5              & \cmark & $59.3$              & $52.8$         & -           \\
                OmniVec              & OmniVec-4           & \cmark & $60.8$              & -           & -           \\
                \shline
                GeminiFusion         & MiT-B$5$ (MiT-B$2$) & \xmark & $57.7$              & $53.3$      & $\bf{66.9}$ \\
                GeminiFusion         & Swin-Large-22k      & \xmark & $60.2$              & $\bf{54.6}$ & -           \\
                GeminiFusion$^*$     & Swin-Large-22k      & \cmark & $\bf{60.9}$         & -           & -           \\
                \shline
            \end{tabular}
}
\end{table}
Upon completion of the four encoding stages, we obtain $M$ sets of feature maps at different stages, denoted as $\boldsymbol{f}_l^m\in\{\boldsymbol{f}_1^m, \boldsymbol{f}_2^m, \boldsymbol{f}_3^m, \boldsymbol{f}_4^m\}$ for each modality $m\in[0, M-1]$. For the $l$-{th} encoding stage, the number of blocks per branch is specified by $b_l \in \{4, 8, 16, 32\}$, the stride by $s_l \in \{4, 8, 16, 32\}$, and the channel dimension by $C_l \in \{64, 128, 320, 512\}$. Within each stage, the $M$ feature maps are fused into a singular feature map $\boldsymbol{f}$ through a process of weighted summation.
Following the encoding process, the resultant four-stage features $\boldsymbol{f}_l \in \{\boldsymbol{f}_1, \boldsymbol{f}_2, \boldsymbol{f}_3, \boldsymbol{f}_4\}$ are channeled into the decoder. The decoder is responsible for synthesizing the segmentation predictions. We employ an MLP-based decoder, as outlined in SegFormer~\cite{segformer}, to serve as our segmentation head.

By employing a single-branch design, we not only streamline network complexity but also enhance predictive generalization capabilities. Moreover, the shared parameter strategy aids in the detection of common patterns across different modalities, which is a key objective of multimodal fusion. It should be noted that while our method excels in processing homogeneous modalities where each data type represents a different perspective of the same input, it currently does not accommodate heterogeneous data combinations, such as images paired with audio or text. We also need to pre-define the method for aligning with the above data pairs. Addressing this limitation remains an avenue for future research.

\section{Experiment}
\subsection{Datasets}
For multimodal semantic segmentation experiments, we use the following datasets:
\textbf{NYUDv2}~\cite{nyu} dataset provides $795$ training and $654$ testing images, labeled into $40$ categories. The resolution we use is $480$x$640$, which is aligned with the setting in CMNeXt~\cite{deliver} and DFormer~\cite{yin2023dformer}.
\textbf{DeLiVER}~\cite{deliver} dataset contains 3983 training and 2005 testing images, which is more than four times the size of the NYUDv2 dataset. It has $25$ classes. The resolution we use is $1024$x$1024$, which is also aligned with CMNeXt. According to CMNext, only mIoU is reported. Thus, we also only report mIoU in experiments on the DeLiVER dataset.
\textbf{SUN RGB-D}~\cite{song2015sun} dataset contains $5285$ training and $5050$ testing images, which is about seven times the size of the NYUDv2 dataset and 1.7 times the size of the DeLiVER dataset. The input resolution is $480$x$480$, which is aligned with DFormer. The class number of the SUN RGB-D dataset is $37$.

For the image-to-image translation task, we follow the experiment settings used in CEN~\cite{wang2020deep} and TokenFusion~\cite{tokenfusion}. \textbf{Taskonomy}~\cite{zamir2018taskonomy} dataset is a large-scale indoor scene dataset, which contains about 4 million indoor images. More than 10 modals are provided with each image, like depth, normal, shade, texture and edge. Each modal is of size $512$x$512$. We use the same sampling strategy with CEN and TokenFusion, which takes 1000 training and 500 testing images. Our implementation details can be found in the appendix~\ref{appendix:implement}.

For the $3$D object detection task, we follow the experiment settings used in MVX-Net~\cite{sindagi2019mvx}.
\textbf{KITTI 3D object detection}~\cite{kitti} dataset contains $7481$ training samples and $7518$ test samples. The test difficulty is categorized into three levels: easy, medium and hard, which is based on the size of the object, the degree of visibility (occlusion), and the degree of truncation. In this paper, like MVX-Net~\cite{sindagi2019mvx}, the training set is further split into a training set and a validation set. After splitting, the training set consists of $3712$ samples and the validation set consists of $3769$ samples.

\begin{table}
\newcommand{\blue}[1]{\textcolor{blue}{#1}}
\definecolor{deepgreen}{rgb}{0.07, 0.53, 0.03}
\centering
\setlength{\tabcolsep}{4pt}
\renewcommand{\arraystretch}{1.0}
\caption{\small{Ablation about the relation discriminator on the NYUDv2 dataset. All training epoch numbers are aligned. We use the MiT-B$3$ as the backbone.}
\vspace{-4pt}
}
\label{tab:ab_relation_discriminator}
\resizebox{1.0\linewidth}{!}
{
\begin{tabular}{l|ccc}
\shline
Relation Discriminator    &  Pixel Acc. &  mAcc. & mIoU \\
\shline
2layer-MLP & $79.3$ & $69.1$ & $55.7$ \\
2layer-MLP + Sigmoid & $79.5$ & $69.7$ & $55.9$ \\
\textbf{2layer-MLP + Softmax} & $\bf{79.9}$ & $\bf{69.9}$ & $\bf{56.8}$ \\
1x1CNN + Softmax & $79.2$ & $69.2$ & $55.7$ \\
3x3CNN + 1x1CNN + Softmax& $79.4$ & $69.6$ & $55.6$ \\
5x5CNN + 3x3CNN + 1x1CNN + Softmax & $79.1$ & $68.7$ & $54.9$ \\
5x5CNN + 3x3CNN + 1x1CNN + 2layer-MLP + Softmax & $79.2$ & $68.9$ & $55.3$ \\
\shline
\end{tabular}
}
\end{table}

\begin{table}
\newcommand{\blue}[1]{\textcolor{blue}{#1}}
\definecolor{deepgreen}{rgb}{0.07, 0.53, 0.03}
\small
\centering
\setlength{\tabcolsep}{3pt}
\renewcommand{\arraystretch}{1.1}
\caption{\small{Ablation about the noise selection on the NYUDv2 dataset. All training epoch numbers are aligned. We use the MiT-B$3$ backbone.}
\vspace{-4pt}
}
\label{tab:ab_noise}
\begin{tabular}{l|ccc}
\shline
Noise type    &  Pixel Acc. &  mAcc. & mIoU \\
\shline
Random Gaussian Noise, Multiply  & $79.2$ & $69.3$ & $55.5$ \\
Random Gaussian Noise, Add & $79.2$ & $68.8$ & $55.3$ \\
Learnable parameter, Multiply & $79.6$ & $69.2$ & $56.2$ \\
\textbf{Learnable parameter, Add} & $\bf{79.9}$ & $\bf{69.9}$ & $\bf{56.8}$ \\
\shline
\end{tabular}
\end{table}


\subsection{Comparisons with TokenFusion}
Table~\ref{tab:tf_ge_seg} summarizes the comparative analysis between GeminiFusion and TokenFusion on segmentation tasks. Overall, with consistent training and testing conditions, GeminiFusion outperforms TokenFusion across the board when it comes to the fusion of two to four modalities. Specifically, in scenarios where RGB is fused with Depth, GeminiFusion achieves an improvement of approximately 1\%-2.6\% over TokenFusion. When all four modalities are fused, GeminiFusion further extends its lead by a significant margin of 3.4\% in mIoU, underscoring the efficacy of our attention-based fusion approach that retains essential information without loss.

Table~\ref{tab:tf_ge_img2img} presents the corresponding results for the image-to-image translation task. Our GeminiFusion outstrips TokenFusion across all evaluated settings. For instance, in the Shade+Texture$\rightarrow$RGB task, GeminiFusion attains FID/KID scores of 41.32/0.81, which is notably superior to TokenFusion with a relative decrease of 12.6\% in the FID metric. Qualitative results, as illustrated in Figure~\ref{fig:vis_img2img}, reveal that predictions using our GeminiFusion exhibit more natural patterns and are smoother and clearer in terms of colors and details. This demonstrates GeminiFusion's capability to preserve a more complete spectrum of the shade information.

\begin{table}
\newcommand{\blue}[1]{\textcolor{blue}{#1}}
\definecolor{deepgreen}{rgb}{0.07, 0.53, 0.03}
\centering
\setlength{\tabcolsep}{3pt}
\renewcommand{\arraystretch}{1.1}
\footnotesize
\caption{\small{Multimodal semantic segmentation results on NYUDv2 and SUN RGB-D datasets by adding our GeminiFusion only to last $k$ layers. All models use the MiT-B$3$ backbone. All training epoch numbers are aligned. Latency is measured by averaging all validation samples in the NYUDv2 dataset.}
}
\label{tab:last_k_semantic_seg}
\resizebox{1.0\linewidth}{!}
{
\begin{tabular}{l|c|ccc|c|c}
\shline
Method     & $k$ &  Param(M) &  GFLOPs & Latency(ms) & \makecell[c]{NYUDv2 \\ mIoU}& \makecell[c]{SUN RGB-D \\mIoU} \\
\shline
TokenFusion     & $28$     & $45.9$ & $108$ & $126$ & $54.2$ & $51.4$\\
\shline
GeminiFusion       & $28$     & $75.8$ & $174$ & $153$ & $\bf{56.8}$ & $\bf{52.7}$\\
GeminiFusion       & $22$     & $75.1$ & $165$ & $144$ & $\bf{56.5}$ & $\bf{52.5}$\\
GeminiFusion       & $16$     & $69.3$ & $152$ & $129$ & $\bf{56.4}$ & $\bf{52.5}$\\
GeminiFusion       & $10$     & $62.5$ & $138$ & $\bf{116}$ & $\bf{56.4}$ & $\bf{52.2}$\\
GeminiFusion       & $4$     & $55.7$ & $124$ & $\bf{103}$ &  $\bf{56.1}$ & $\bf{51.9}$\\
GeminiFusion       & $1$     & $48.8$ & $119$ & $\bf{102}$ &  $\bf{55.1}$ & $\bf{51.9}$\\
GeminiFusion       & $0$     & $45.9$ & $108$ & $\bf{95}$ &  $53.3$ & $51.2$\\
\shline
\end{tabular}
}
\end{table}

\begin{table}
\begin{minipage}[t]{1\linewidth}
\small
\centering
\setlength{\tabcolsep}{8pt}
\renewcommand{\arraystretch}{1.0}
\caption{\small{Ablation about different parts of GeminiFusion on the NYUDv2 dataset. PWC: point-wise cross-attention, NSA: noised self-attention, ARD: attention relation discriminator.}}
\label{tab:abl_method}
\begin{tabular}{c|c|c|c}
\shline
PWC &  NSA & ARD    &  mIoU\\
\shline
\xmark       & \xmark & \xmark  & $53.3$            \\
\cmark       & \xmark & \xmark  & $55.4$            \\
\cmark       & \cmark & \xmark  & $56.3$            \\
\cmark       & \cmark & \cmark  & $\bf{56.8}$       \\
\shline
\end{tabular}
\end{minipage}
\end{table}

\subsection{Applying to Swin Transformer}
The proposed GeminiFusion module is a plug-and-play module that can be inserted into existing multimodal architectures (predominantly into encoders) for enhancing the model's cross-modal learning capabilities. This modular approach allows GeminiFusion to take advantage of different architectures to improve the model's performance in multimodal tasks. In the previous experiments, we follow the TokenFusion codebase, which uses the SegFormer~\cite{segformer} as the encoder and a simple FFN as the decoder. However, in addition to SegFormer, models such as Swin Transformer~\cite{swin} can also be used as encoder models, which together with the decoder form a complete segmentation model. We further conducts several experiments on the Swin Transformer. Specifically, we inserts GeminiFusion into the SwinBlock. The official checkpoints of Swin Transformer pre-trained on the ImageNet classification task can also be loaded directly without degradation of accuracy, which demonstrates the advantages of our approach. The experimental results are shown in Table~\ref{tab:swin_seg}. It can be seen that GeminiFusion is also applicable in frameworks such as Swin Transformer, and in the case of using the Swin-Large-22k model, which was pre-trained on a larger ImageNet-22k dataset and with a larger number of parameters, as the baseline model, GeminiFusion also achieves optimal results among encoders, which reflects the plug-and-play nature of GeminiFusion in different frameworks, as well as its ability to successfully leverage the better representational capabilities provided by larger encoders.

\subsection{Comparisons with state-of-the-art methods}

In this paper, GeminiFusion is benchmarked against state-of-the-art multimodal segmentation methods on NYUDv2, SUN RGB-D, and DeLiVER datasets, and the results are detailed in Table~\ref{tab:sota_results}. To ensure the fairness of the comparison, all methods that use pre-training methods and training strategies other than pre-training on the ImageNet classification tasks are labeled as “Additional
Strategies”, such as PolyMax~\cite{yang2024polymax} (pre-training is performed using ImageNet-22K and Taskonomy), DPLNet~\cite{dong2023efficient} (using pre-trained segmentation model), OmniVec~\cite{srivastava2024omnivec} (pre-trained based on self-supervision of large-scale masks), DFormer~\cite{yin2023dformer} (utilizes an RGB-D pre-trained backbone), EMSANet~\cite{seichter2023panopticndt} and OMNIVORE~\cite{girdhar2022omnivore} (both of which utilize a strategy of multi dataset pre-training coupled with fine-tuning of individual datasets). In particular, we likewise attempts an additional pre-training strategy, using a GeminiFusion model (Swin Large-22k backbone) trained on the SUN RGBD dataset and fine-tuned on the NYUDv2 dataset.

As can be seen from the experimental results, GeminiFusion using the Swin-Large-22k backbone network achieves the highest level of performance on both NYUDv2 and SUN RGB-D datasets.
Moreover, when fusing modalities such as RGB with Depth, Event and LiDAR data, GeminiFusion with the MiT-B$2$ backbone secures substantial gains over CMNeXt, attesting to the efficacy of our pixel-wise fusion methodology in handling highly aligned modalities. Additionally, we juxtapose the performance of the MiT-B4-based GeminiFusion with CMNeXt on the NYUDv2 dataset, as illustrated in Figure~\ref{fig:latency_miou}. Here, GeminiFusion not only attains marginally superior results but also boasts significantly reduced latency, even in the absence of multi-scale and flip testing augmentations typically employed by CMNeXt. 

\subsection{Effect of each component on GeminiFusion}
We present an ablation study on the NYUDv2 dataset to assess the contribution of each component within our GeminiFusion framework. Table~\ref{tab:abl_method} shows our implementation of point-wise cross-attention yields a 2.1\% increase in mIoU compared to the baseline, demonstrating that direct information exchange between modalities can lead to substantial gains. Additionally, the effectiveness of the noise-adaptive self-attention mechanism is evidenced by its ability to preserve intra-modal features, thereby preventing the loss of valuable information. The proposed relation discriminator can help refine the generation process of key features within the attention mechanism, ensuring more precise adjustments that improve overall performance.

\subsection{Discussion on Inference Latency}
Contrary to the TokenFusion approach as documented in~\cite{tokenfusion}, our GeminiFusion method does not require integration at every layer within the network architecture. As evidenced by the experiments in Table~\ref{tab:last_k_semantic_seg}, implementing GeminiFusion in only the final 10 layers still yields faster inference speeds while preserving accuracy, outperforming the benchmark method. Incorporating GeminiFusion even in just the last layer alone surpasses TokenFusion in terms of both inference latency and accuracy. However, it should be noted that the optimal results are achieved when GeminiFusion is applied across every layer.

Figure~\ref{fig:latency_miou} graphically represents the trade-off between performance and latency. The comparison clearly demonstrates that our GeminiFusion significantly outperforms TokenFusion in terms of efficiency by a considerable margin.

\subsection{3D Object Detection task}
We choose the MVX-Net~\cite{sindagi2019mvx} framework and the KITTI dataset for our $3$D object detection experiments for vehicles.
The experiments use images and depth maps as inputs for the detection of vehicle categories in the KITTI dataset, which is aligned with other works~\cite{zhang2023glenet,zheng2021se}. For the processing of the KITTI dataset, we choose the same dataset division and data processing methods as MVX-Net. GeminiFusion is inserted into the original fusion layer of MVX-Net, and the experimental results are shown in Table~\ref{tab:3d_det}, which show that GeminiFusion achieves significant improvement in most of the performance indexes with almost no increase in the number of parameters, and a few performance indexes are almost the same as the benchmark model.


\section{Conclusion}
In this paper, we comprehensively examine exchange-based cross-modal transformers and point out their intrinsic deficiency in achieving comparable performance of cross-attention mechanisms. Furthermore, we propose a pixel-wise fusion approach named GeminiFusion, combining intra-modality and inter-modality attention for dynamic integration of complementary information across modalities. GeminiFusion achieves state-of-the-art performance across various multimodal semantic segmentation benchmark datasets, and also proved its effectiveness on image-to-image translation and $3$D object detection tasks. It is worth noting that GeminiFusion operates with linear complexity with respect to the number of input tokens, achieving efficiency comparable with unimodal counterparts.

\section*{Acknowledgements}
Ding Jia and Chao Zhang are supported by the National Nature Science Foundation of China under Grant 62071013 and 61671027, and National Key R\&D Program of China under Grant 2018AAA0100300. Chang Xu is supported in part by the Australian Research Council under Projects DP240101848 and FT230100549.

\section*{Impact Statement}
This paper contributes to the advancement of multimodal feature fusion in Machine Learning by comparing exchange-based fusion with cross-attention based fusion. Our findings consistently demonstrate that cross-attention based fusion outperforms exchange-based fusion by effectively preserving core information among features from different modalities. Additionally, we propose an efficient GenimiFusion approach to reduce the computational overhead associated with cross-attention. There are many potential societal consequences of our work, none which we feel must be specifically highlighted here.


{
\small
\bibliography{egbib}
\bibliographystyle{icml2024}
}

\newpage
\appendix
\onecolumn

\section{Implementation Details}
\label{appendix:implement}
\begin{itemize}
    \item In the context of \textbf{multimodal semantic segmentation}, our training hyper-parameters are developed by following the methodologies from the TokenFusion~\cite{tokenfusion} and CMNeXt~\cite{deliver} codebases. For model training, we employ NVIDIA V100 GPUs in configurations of 3, 4, and 8 units for the NYUDv2, SUN RGB-D, and DeLiVER datasets, respectively, adhering to the same environmental settings as specified in the original papers. Our encoder design is an adaptation from SegFormer~\cite{segformer}, which has been pre-trained solely on the ImageNet-1K~\cite{deng2009imagenet} dataset for classification tasks. For experiments on the NYUDv2 and SUN RGB-D datasets, we utilize the setup from the TokenFusion, maintaining consistency in batch size, optimizer, learning rate, and learning rate scheduler. Within our proposed GeminiFusion model, we configure the number of attention heads to $8$. To mitigate the risk of overfitting, we set the drop path rate to $0.4$, while the drop rate remains at $0.0$. Conversely, for the DeLiVER dataset, our foundation training hyper-parameters are the same with CMNeXt, which necessitates a smaller backbone. Consequently, we reduce the drop path rate to $0.2$. All other parameters, including batch size, optimizer, weight decay, and learning rate scheduler, remain in line with CMNeXt's original configuration, except for the learning rate, which is modified to $2e^{-4}$.
    
    \item For the \textbf{image-to-image translation} task, we also follow the setting in TokenFusion and set the same hyper-parameters as the TokenFusion. We use one NVIDIA V100 card for all image-to-image translation experiments. 
    
    \item For the \textbf{3D object detection} task, we also follow the setting in MVX-Net and set the same hyper-parameters as the MVX-Net. We use 4 NVIDIA V100 cards for all experiments. 
\end{itemize}

\section{More Visualization Results}

\begin{figure}[h]
\qquad\, (a) Input-0 \qquad\quad (b) Input-1 \qquad (c) TokenFusion \qquad\quad (d) Ours \qquad\quad\, (e) GT \\ [1mm]
\centering
\includegraphics[width=26mm]{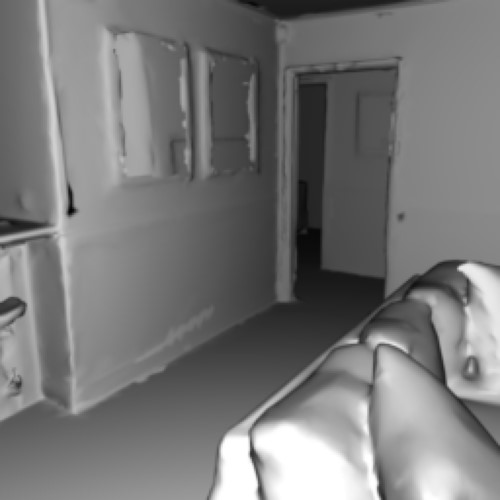}
\includegraphics[width=26mm]{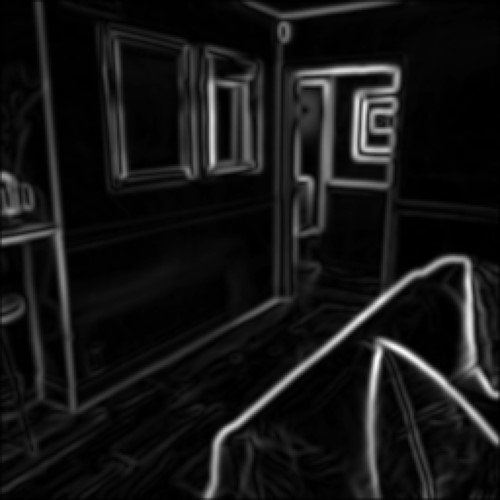}
\includegraphics[width=26mm]{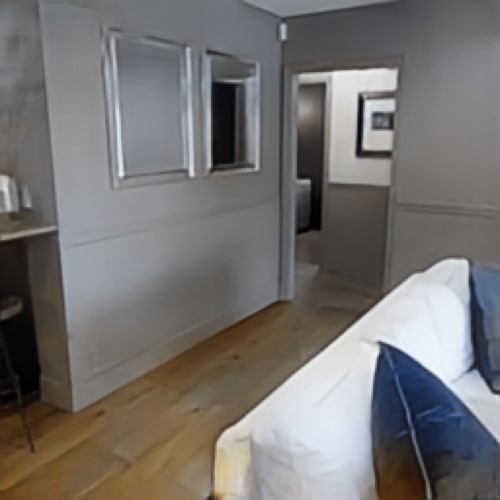}
\includegraphics[width=26mm]{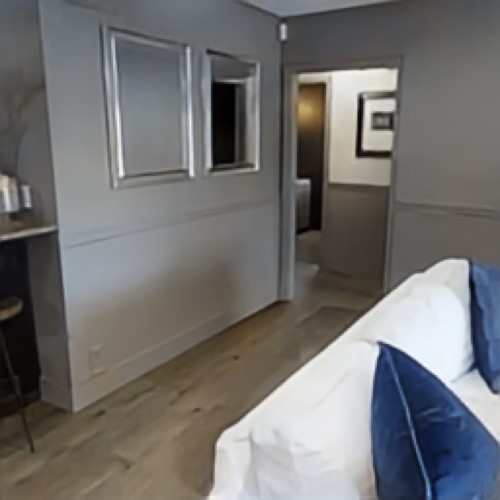}
\includegraphics[width=26mm]{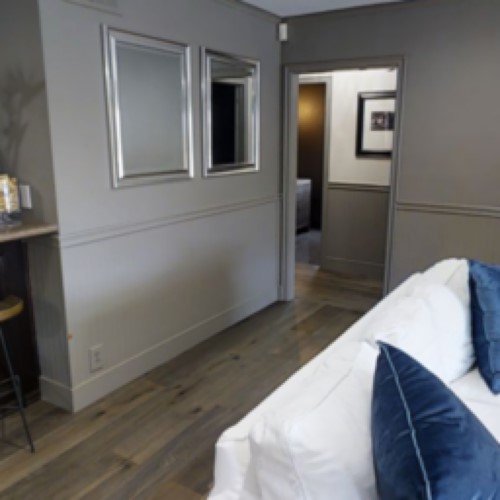} \\

\includegraphics[width=26mm]{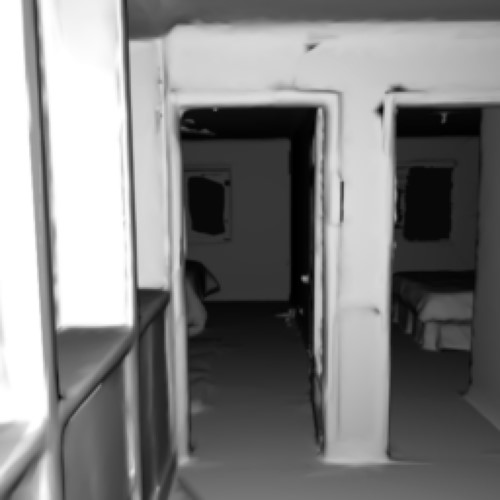}
\includegraphics[width=26mm]{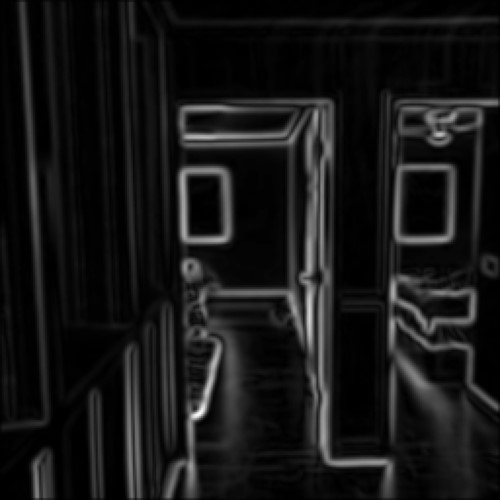}
\includegraphics[width=26mm]{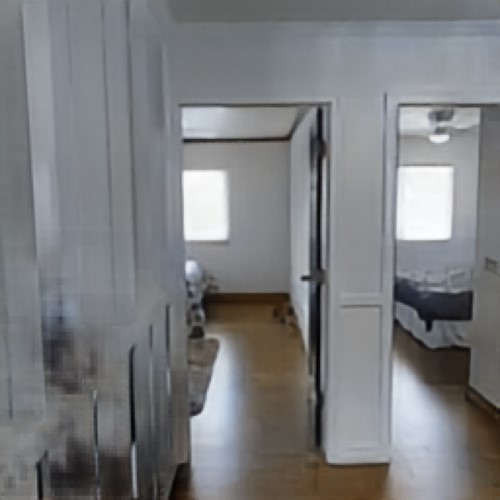}
\includegraphics[width=26mm]{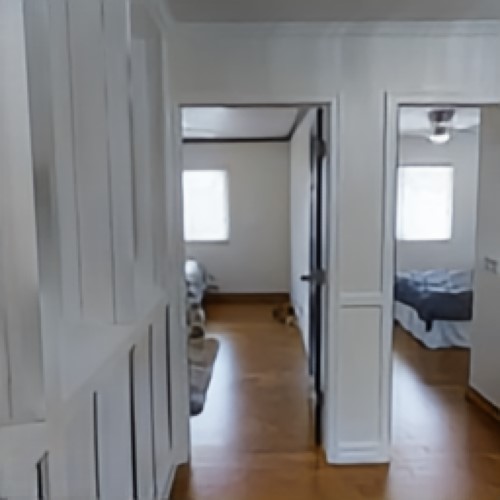}
\includegraphics[width=26mm]{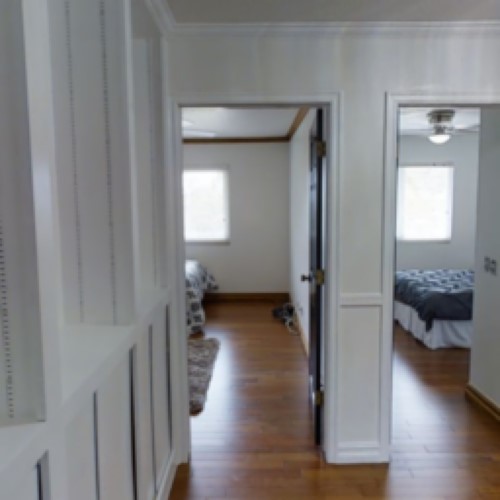}\\

\includegraphics[width=26mm]{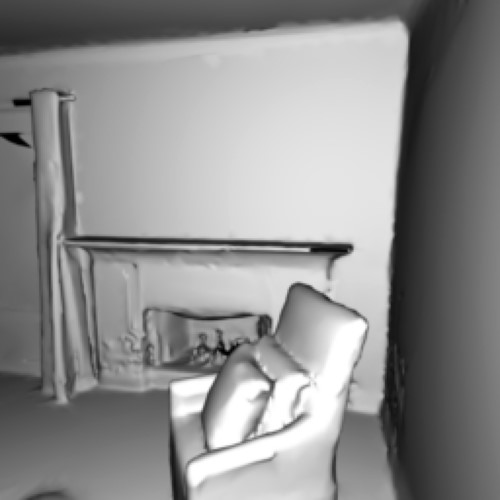}
\includegraphics[width=26mm]{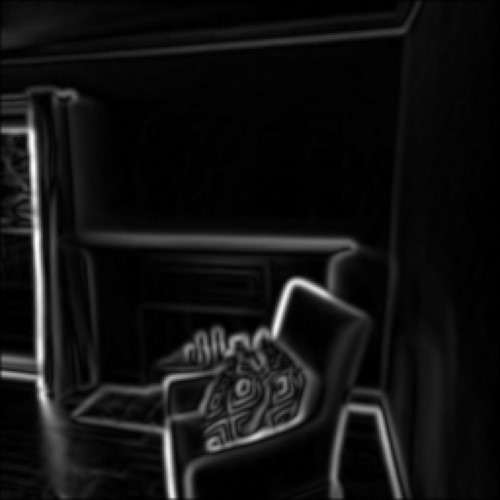}
\includegraphics[width=26mm]{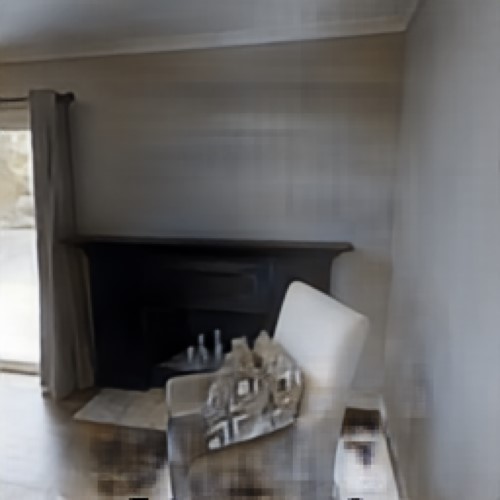}
\includegraphics[width=26mm]{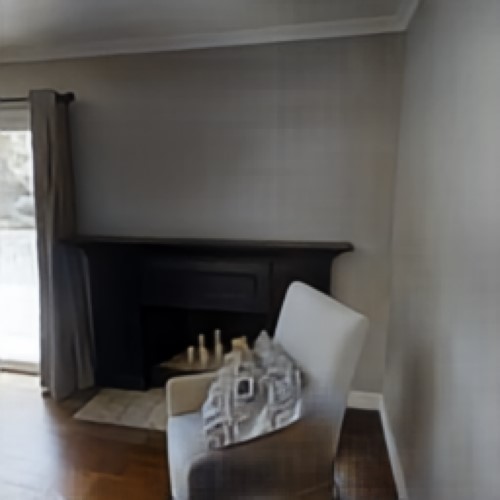}
\includegraphics[width=26mm]{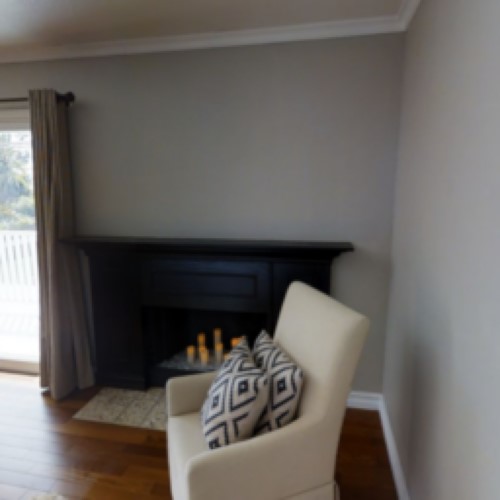}
\caption{\small{Shade+Texture$\rightarrow$RGB. Image-to-image translation results on the validation split of Taskonomy~\cite{zamir2018taskonomy}.}}
\label{fig:ill_img2img_shade_texture_rgb_sup}
\end{figure}

\begin{figure}
\qquad\, (a) Input-0 \qquad\quad (b) Input-1 \qquad (c) TokenFusion \qquad\quad (d) Ours \qquad\quad\, (e) GT \\ [1mm]
\centering
\includegraphics[width=26mm]{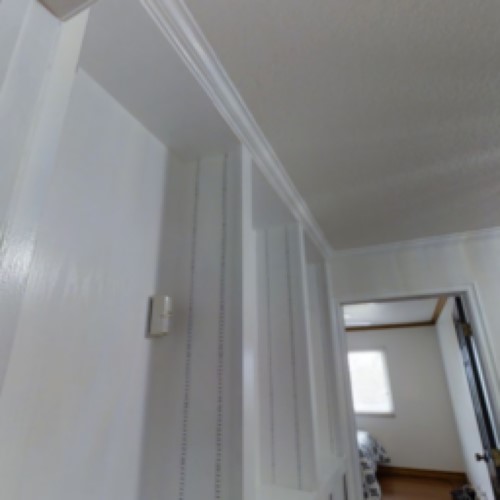}
\includegraphics[width=26mm]{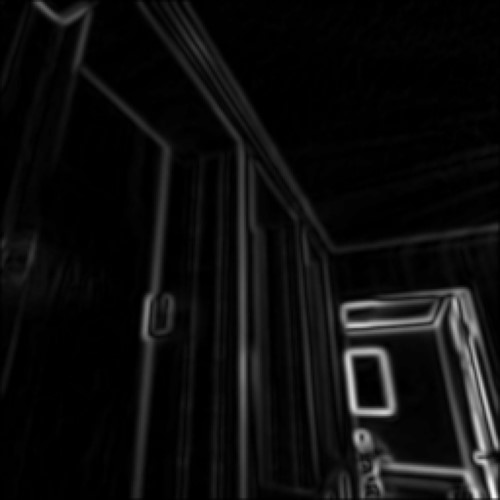}
\includegraphics[width=26mm]{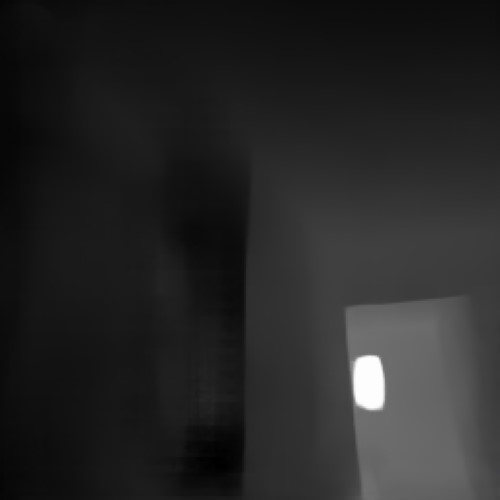}
\includegraphics[width=26mm]{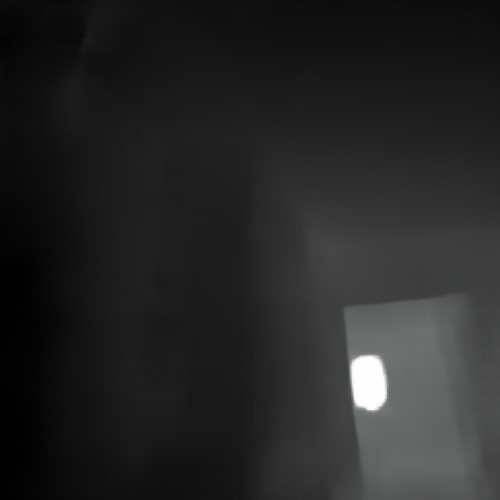}
\includegraphics[width=26mm]{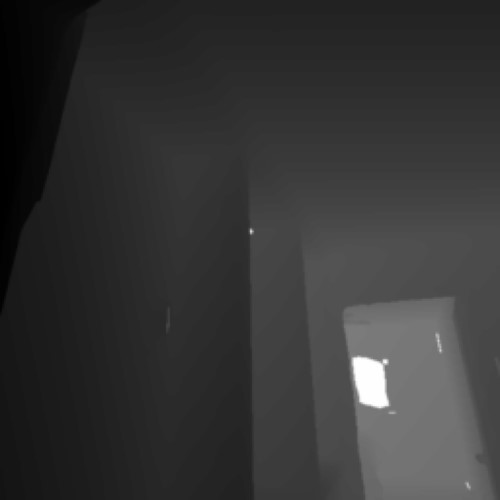} \\

\includegraphics[width=26mm]{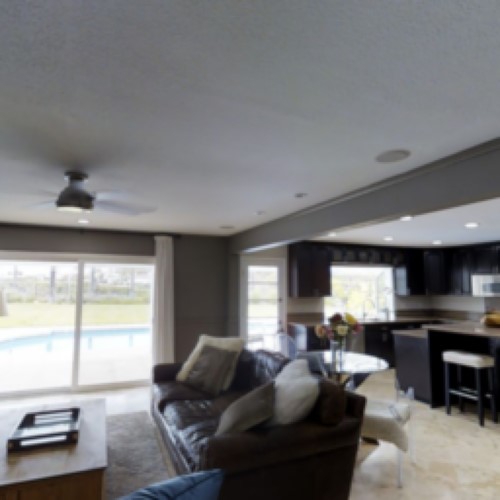}
\includegraphics[width=26mm]{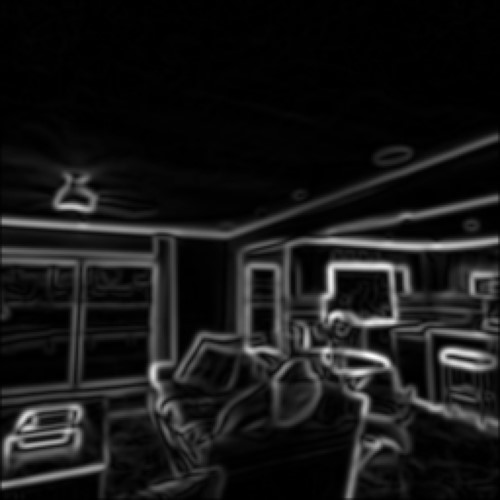}
\includegraphics[width=26mm]{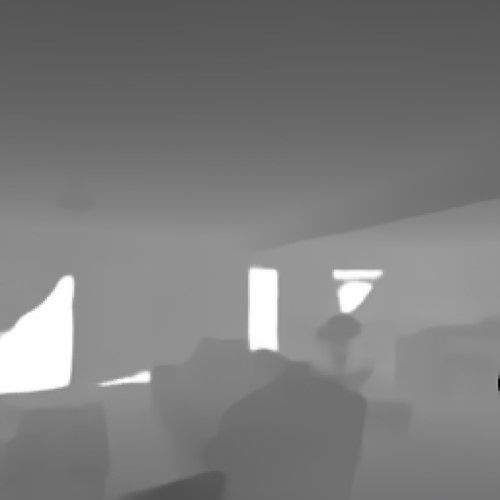}
\includegraphics[width=26mm]{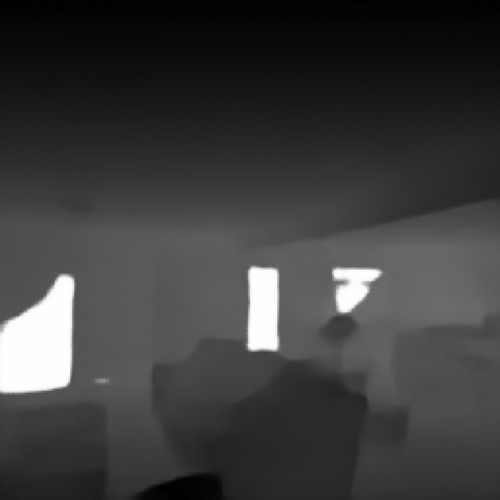}
\includegraphics[width=26mm]{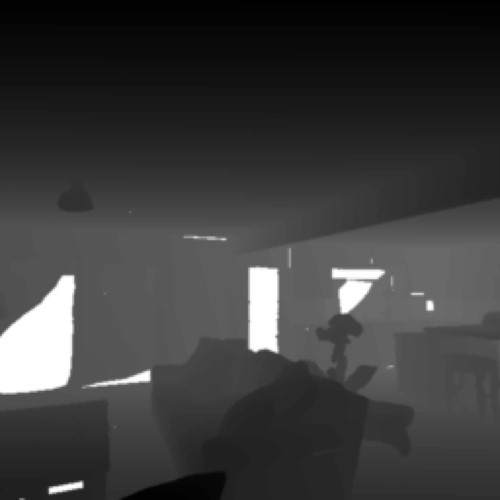} \\

\includegraphics[width=26mm]{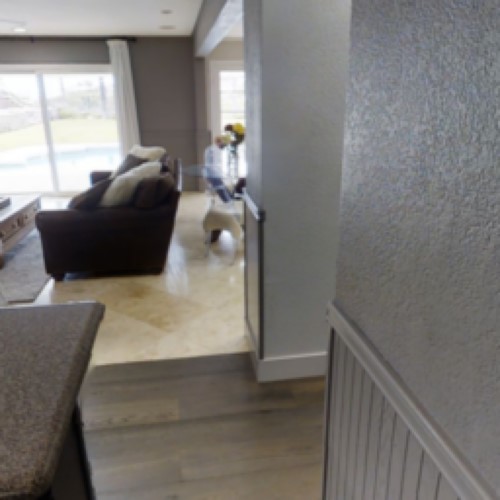}
\includegraphics[width=26mm]{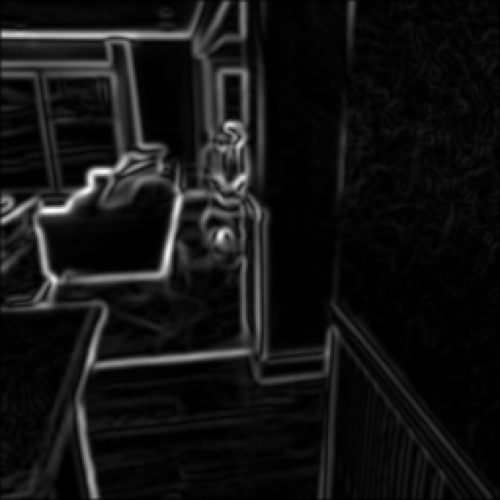}
\includegraphics[width=26mm]{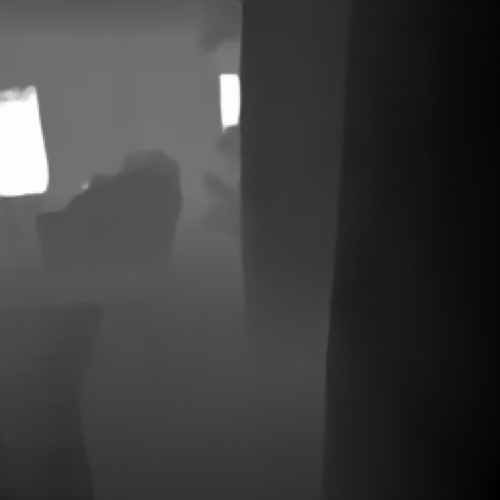}
\includegraphics[width=26mm]{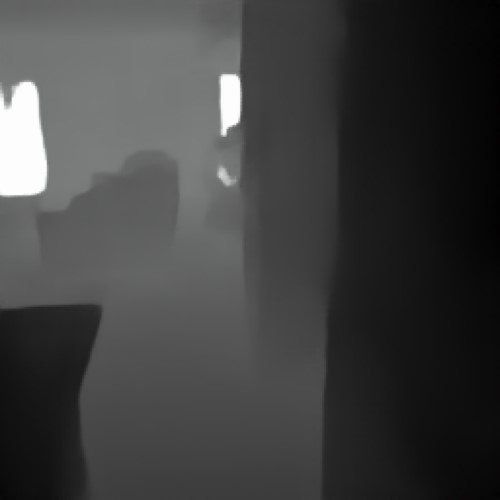}
\includegraphics[width=26mm]{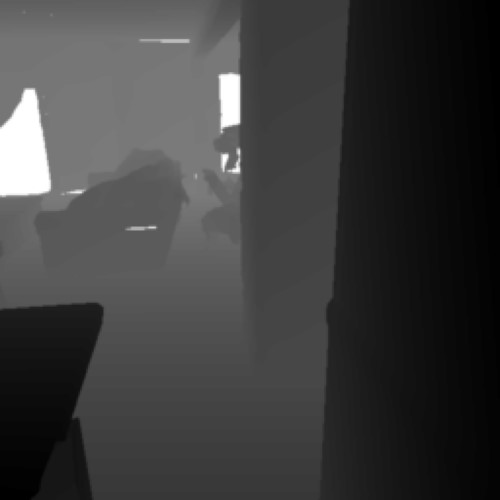}
\caption{\small{RGB+Edge$\rightarrow$Depth. Image-to-image translation results on the validation split of Taskonomy.}}
\label{fig:ill_img2img_rgb_edge_depth_sup}
\end{figure}

\begin{figure}
\qquad\, (a) Input-0 \qquad\quad (b) Input-1 \qquad (c) TokenFusion \qquad\quad (d) Ours \qquad\quad\, (e) GT \\ [1mm]
\centering
\includegraphics[width=26mm]{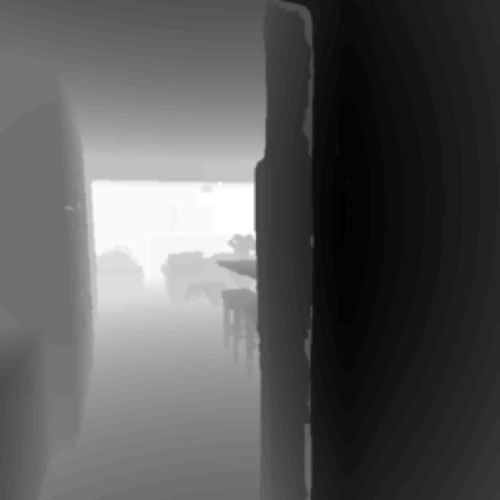}
\includegraphics[width=26mm]{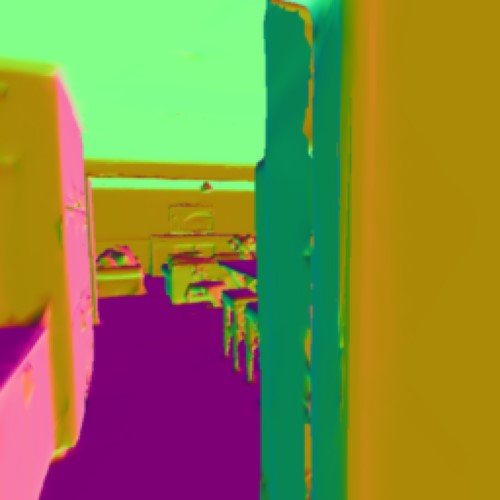}
\includegraphics[width=26mm]{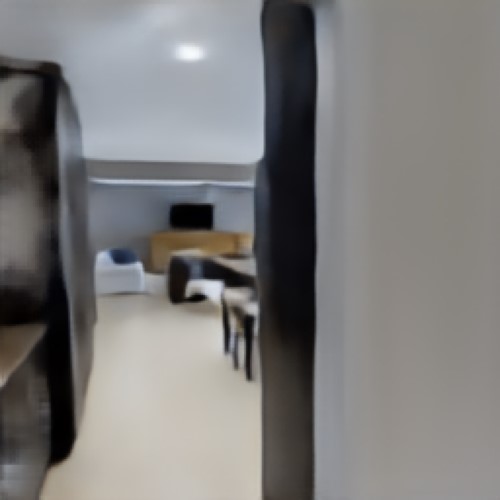}
\includegraphics[width=26mm]{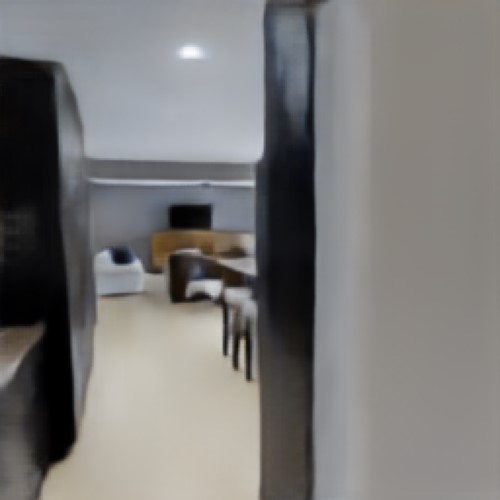}
\includegraphics[width=26mm]{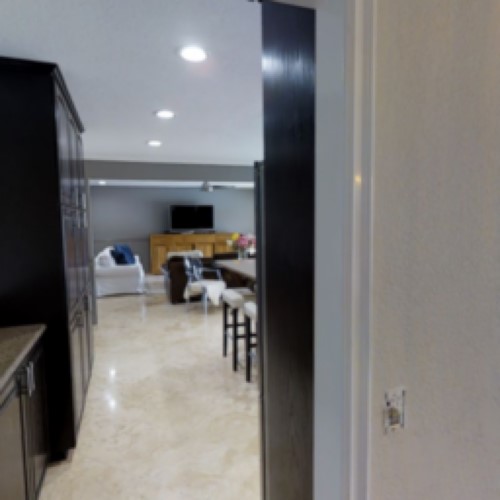} \\

\includegraphics[width=26mm]{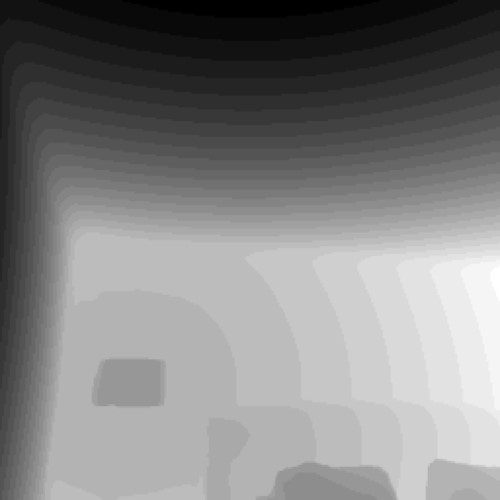}
\includegraphics[width=26mm]{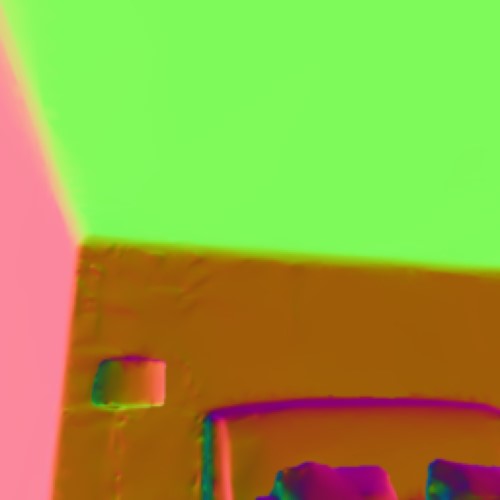}
\includegraphics[width=26mm]{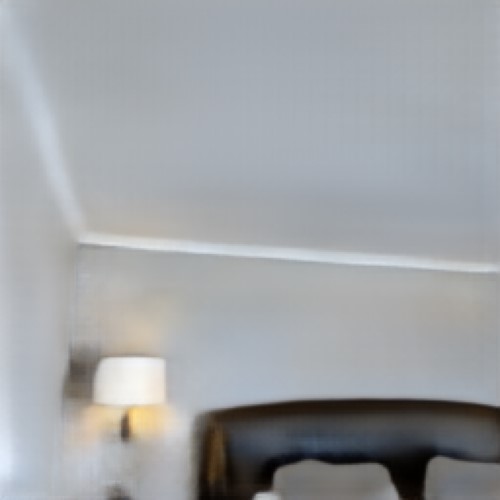}
\includegraphics[width=26mm]{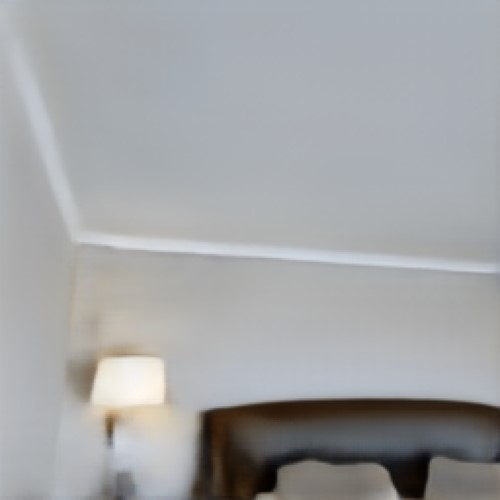}
\includegraphics[width=26mm]{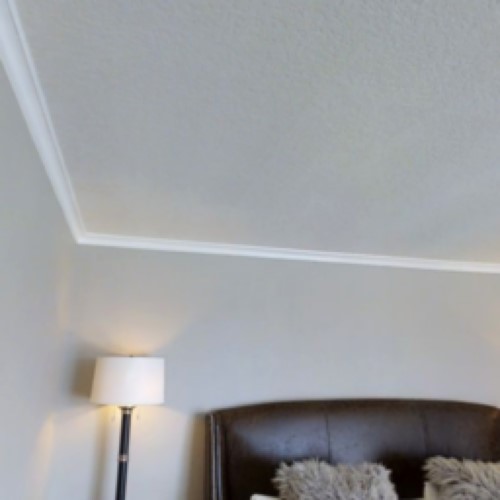} \\

\includegraphics[width=26mm]{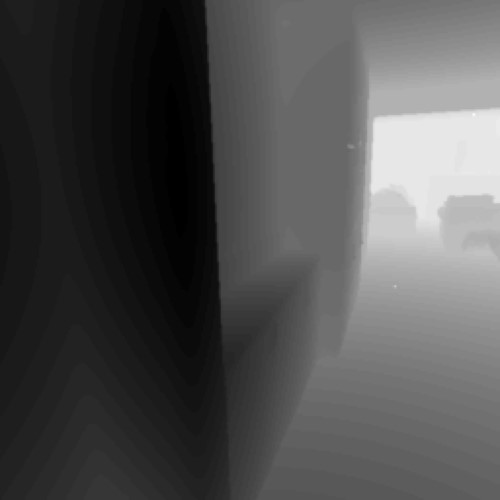}
\includegraphics[width=26mm]{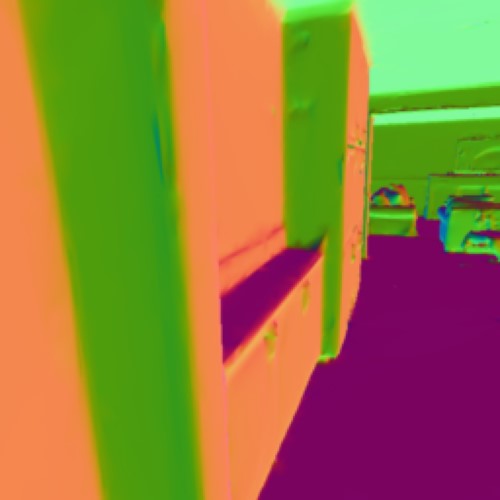}
\includegraphics[width=26mm]{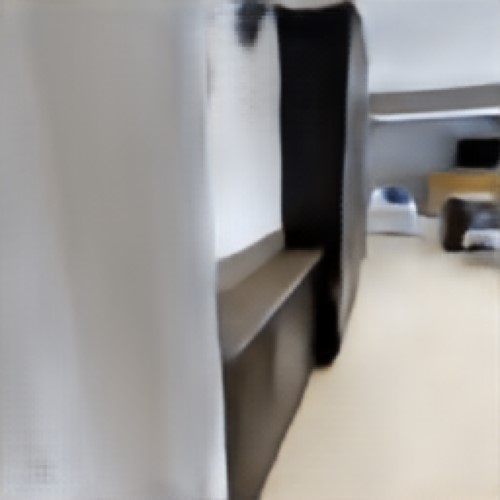}
\includegraphics[width=26mm]{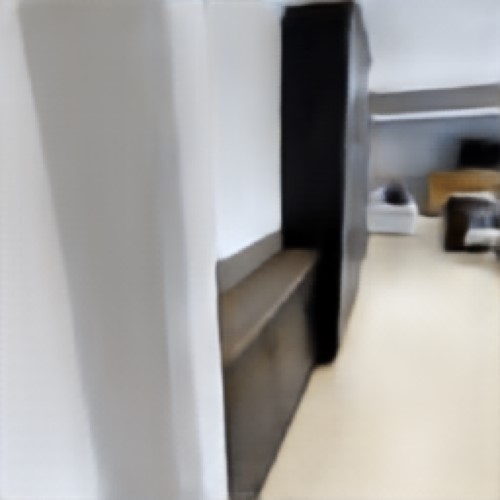}
\includegraphics[width=26mm]{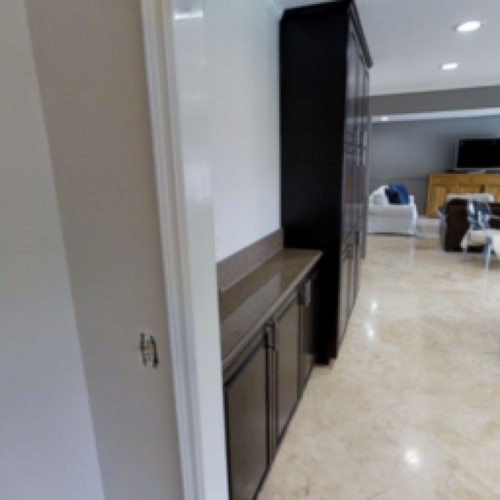}
\caption{\small{Depth+Normal$\rightarrow$RGB. Image-to-image translation results on the validation split of Taskonomy.}}
\label{fig:ill_img2img_depth_normal_rgb_sup}
\end{figure}

\begin{figure}
\qquad\, (a) Input-0 \qquad\quad (b) Input-1 \qquad (c) TokenFusion \qquad\quad (d) Ours \qquad\quad\, (e) GT \\ [1mm]
\centering
\includegraphics[width=26mm]{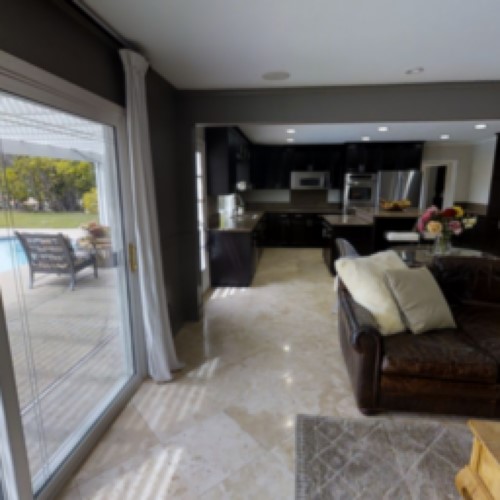}
\includegraphics[width=26mm]{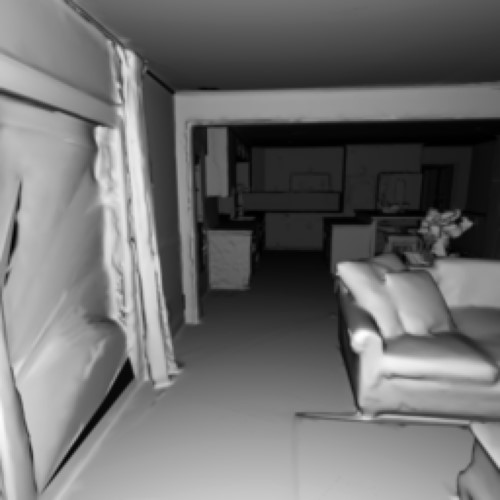}
\includegraphics[width=26mm]{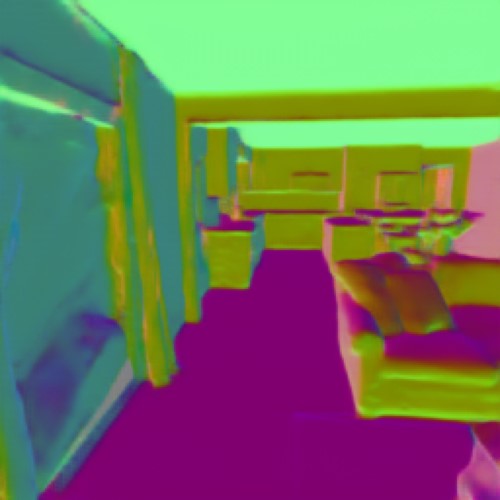}
\includegraphics[width=26mm]{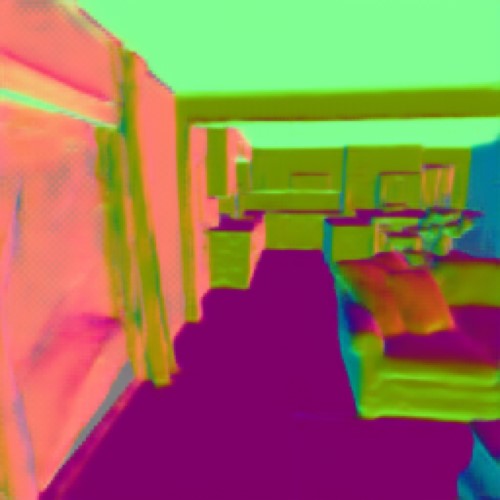}
\includegraphics[width=26mm]{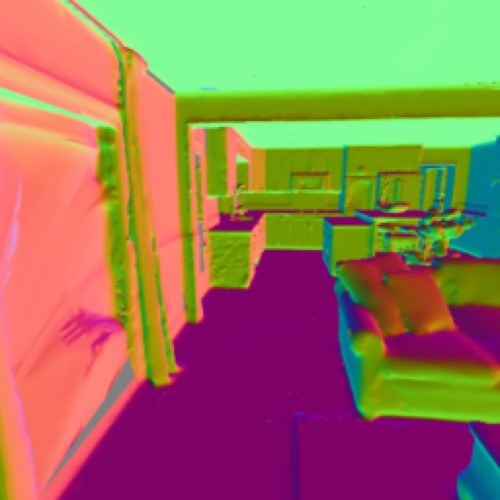} \\

\includegraphics[width=26mm]{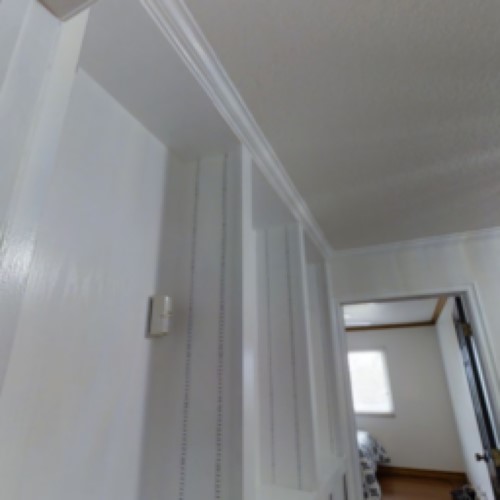}
\includegraphics[width=26mm]{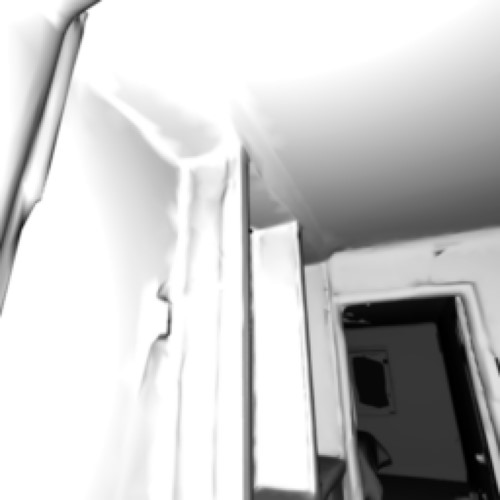}
\includegraphics[width=26mm]{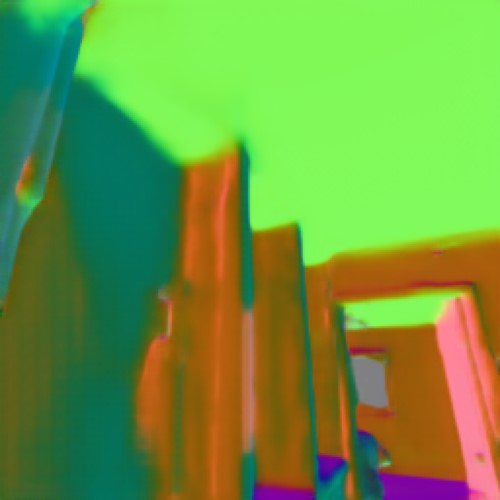}
\includegraphics[width=26mm]{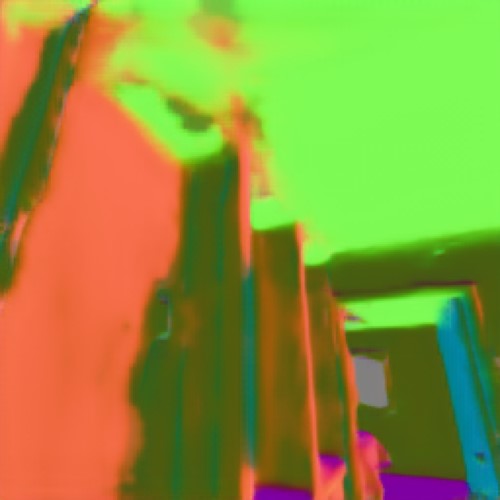}
\includegraphics[width=26mm]{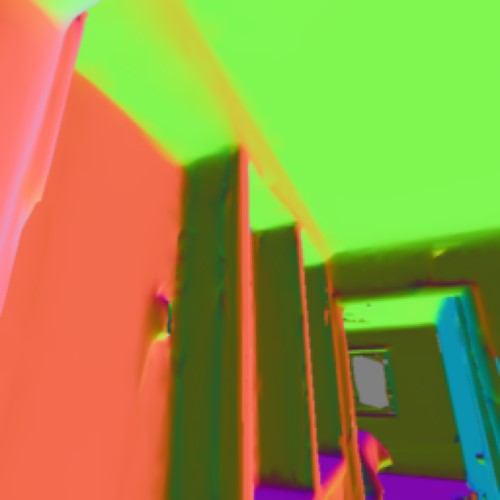} \\

\includegraphics[width=26mm]{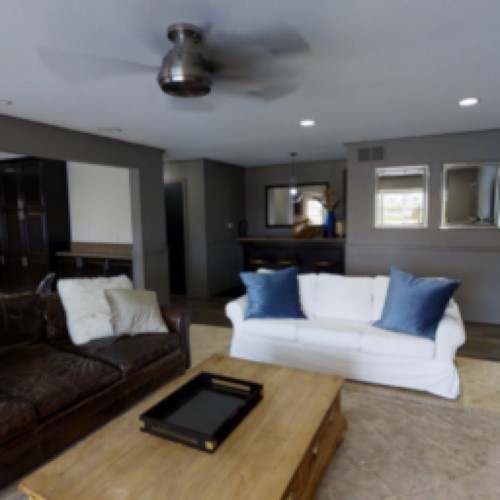}
\includegraphics[width=26mm]{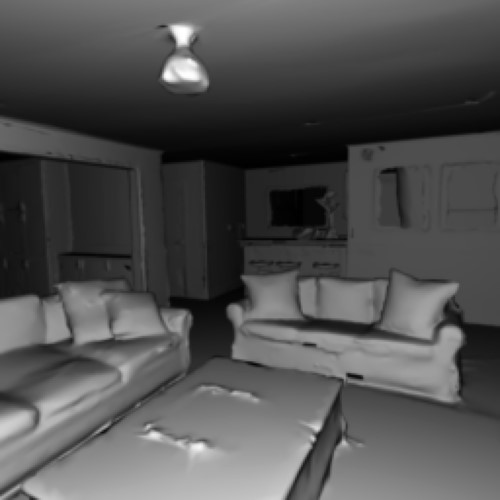}
\includegraphics[width=26mm]{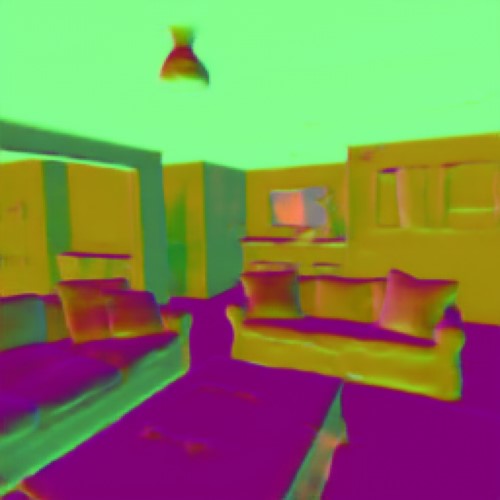}
\includegraphics[width=26mm]{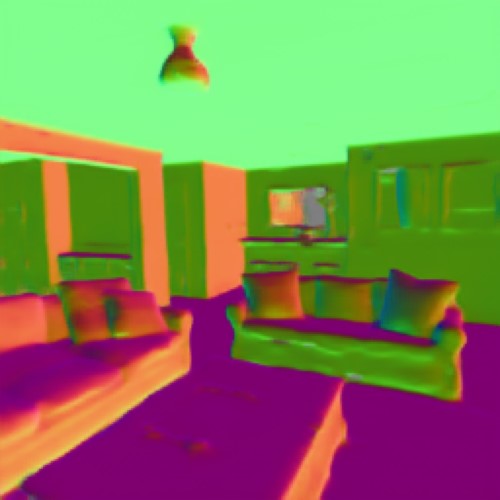}
\includegraphics[width=26mm]{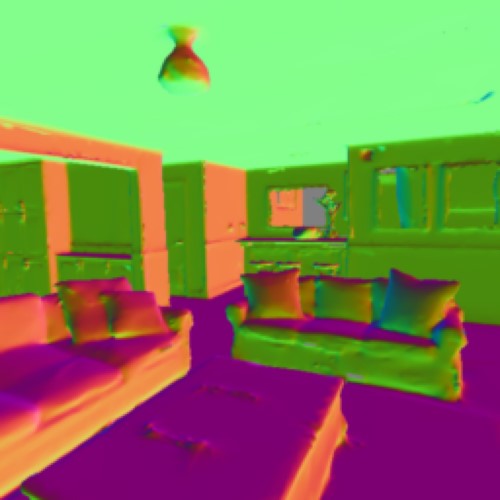}
\caption{\small{RGB+Shade$\rightarrow$Normal. Image-to-image translation results on the validation split of Taskonomy.}}
\label{fig:ill_img2img_rgb_shade_normal_sup}
\end{figure}
\end{document}